\RequirePackage{amsmath,amssymb} 
\documentclass{article}
\usepackage{graphicx}
\usepackage{color}
\usepackage[hratio=1:1,margin=2.3cm]{geometry}

\usepackage[boxed, titlenumbered]{algorithm2e}
\usepackage{color}
\usepackage{subfig}
\usepackage{xifthen}
\usepackage{mathtools}
\usepackage{cite}
\usepackage{forloop}
\usepackage{rotating}

\newcommand{\rulesep}{\unskip\ \vrule\ }

\newcommand{\todo}[1][]{%
\ifthenelse{\isempty{#1}}
	 {\textcolor{red}{(TODO)} \marginpar{\textcolor{red}{$\star$}}}
   {\textcolor{red}{(TODO: \marginpar{\textcolor{red}{$\star$}} #1)}}
}

\newcommand{\vcenteredinclude}[2]{\begingroup
\setbox0=\hbox{\includegraphics[#1]{#2}}%
\parbox{\wd0}{\box0}\endgroup}

\author{
  Calvin Murdock\\
	Machine Learning Dept.\\
	Carnegie Mellon University\\
  \texttt{cmurdock@cs.cmu.edu}
  \and
  Ming-Fang Chang\\
	Robotics Institute\\
	Carnegie Mellon University\\
  \texttt{mingfanc@andrew.cmu.edu}
	\and
  Simon Lucey\\
	Robotics Institute\\
	Carnegie Mellon University\\
  \texttt{slucey@cs.cmu.edu}
}

\date{}
\title{\bf Deep Component Analysis via \\  Alternating Direction Neural Networks}

\begin{document}



\maketitle

\begin{abstract}
Despite a lack of theoretical understanding, deep neural networks have achieved unparalleled performance in a wide range of applications. On the other hand, shallow representation learning with component analysis is associated with rich intuition and theory, but smaller capacity often limits its usefulness. To bridge this gap, we introduce Deep Component Analysis (DeepCA), an expressive multilayer model formulation that enforces hierarchical structure through constraints on latent variables in each layer. For inference, we  propose a differentiable optimization algorithm implemented using recurrent Alternating Direction Neural Networks (ADNNs) that enable parameter learning using standard backpropagation. By interpreting feed-forward networks as single-iteration approximations of inference in our model, we provide both a novel theoretical perspective for understanding them and a practical technique for constraining predictions with prior knowledge. Experimentally, we demonstrate performance improvements on a variety of tasks, including single-image depth prediction with sparse output constraints.

\end{abstract}

\section{Introduction}

\begin{figure}[t]
\begin{minipage}{0.45\textwidth}
\captionsetup[subfigure]{labelformat=empty}
\centering
\subfloat[(a) Feed-Forward]{
\includegraphics[width=0.5\columnwidth]{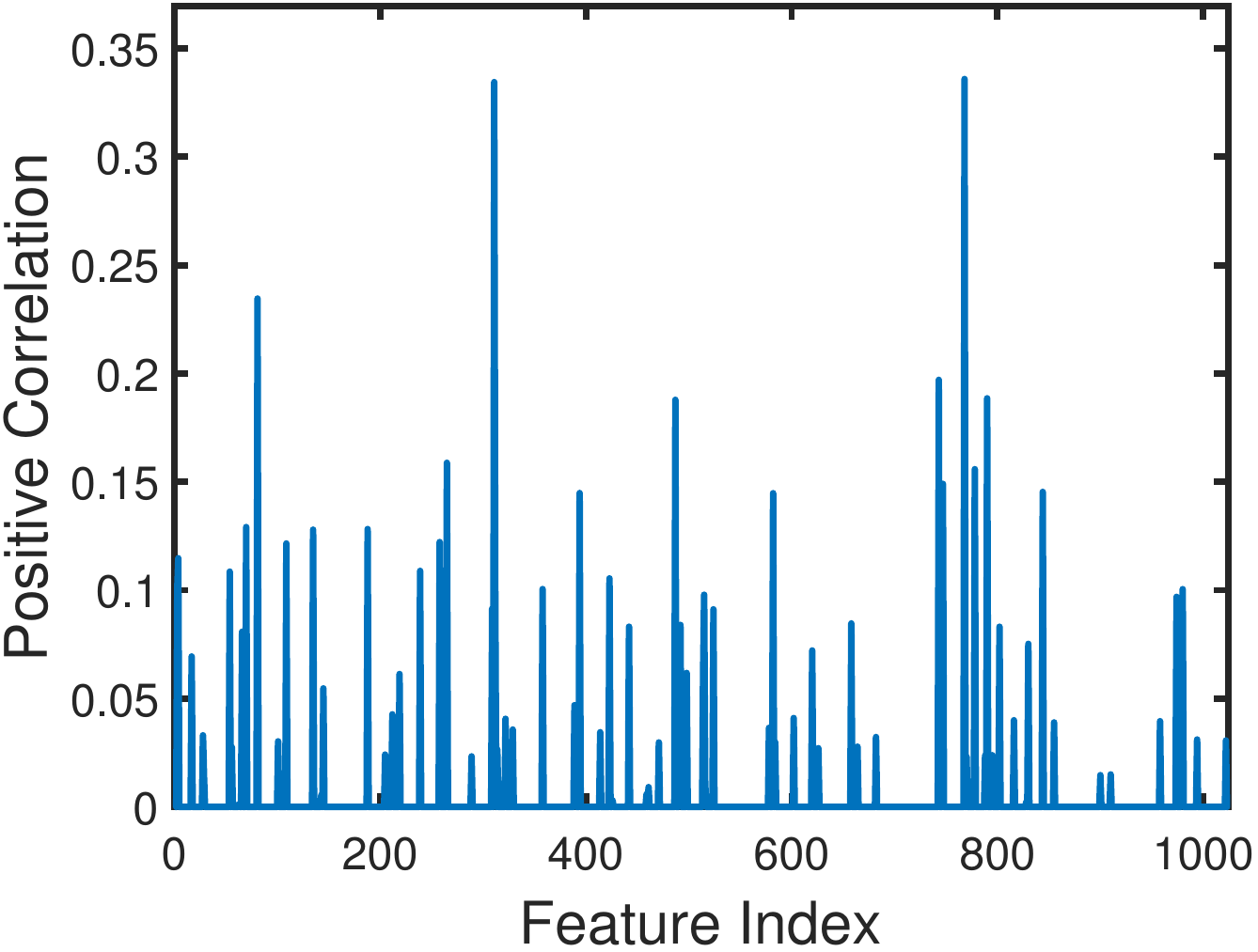}
}
\subfloat[(b) Optimization]{
\includegraphics[width=0.5\columnwidth]{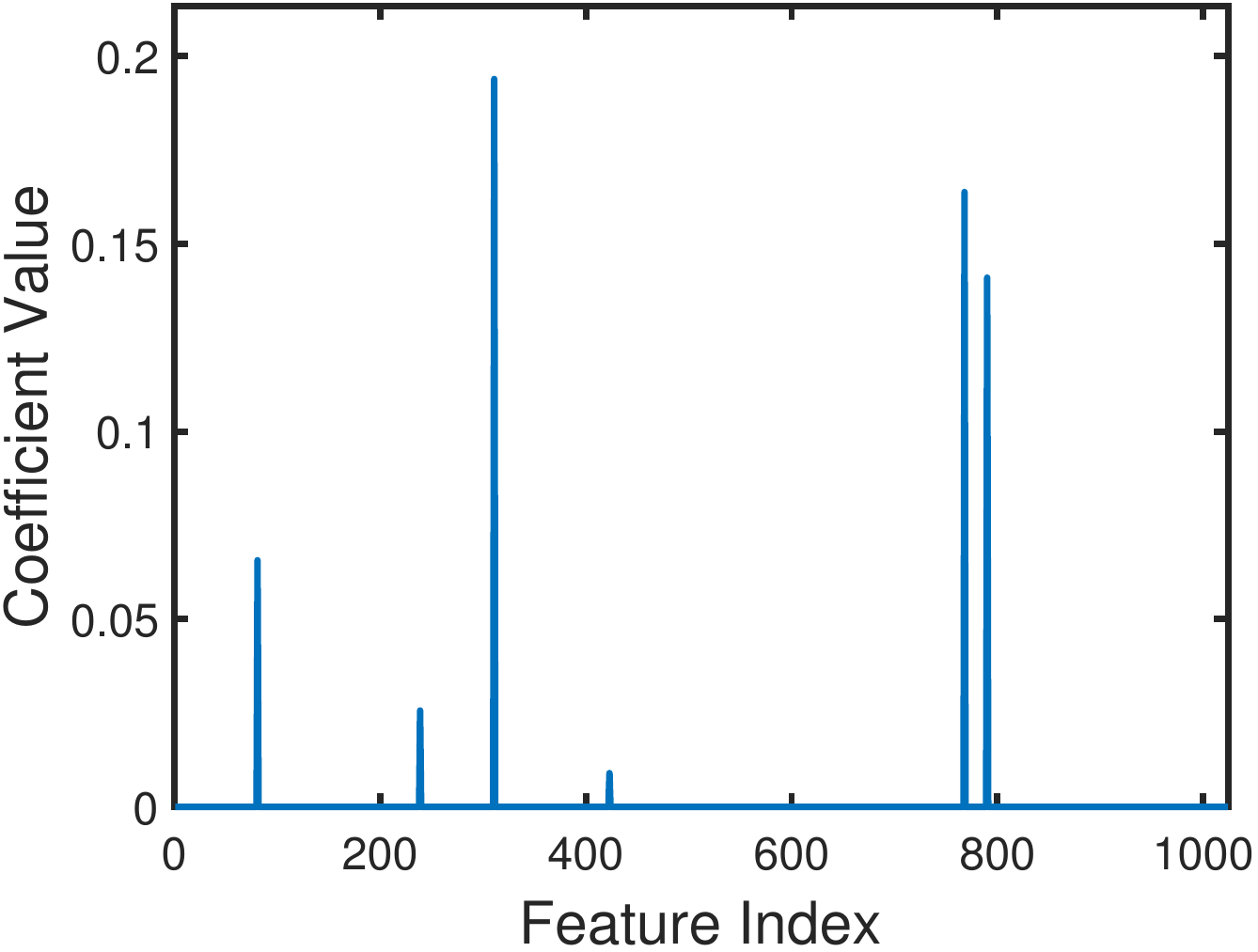}
}
\end{minipage}
\begin{minipage}{0.45\textwidth}
\captionsetup[subfigure]{labelformat=empty}
\vspace{0.3em}
\hspace{0.3em}
\subfloat{\makebox[2.5em]{\footnotesize (c)}}
\subfloat[$\boldsymbol{x}$]{\vcenteredinclude{width=0.115\columnwidth}{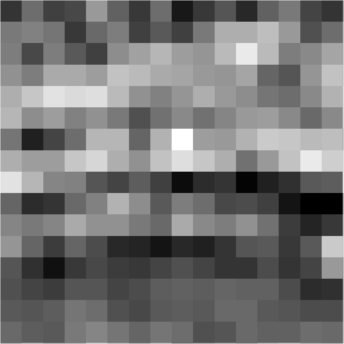}}
\subfloat{\makebox[0.9em]{}}
\subfloat[0.336]{\vcenteredinclude{width=0.115\columnwidth}{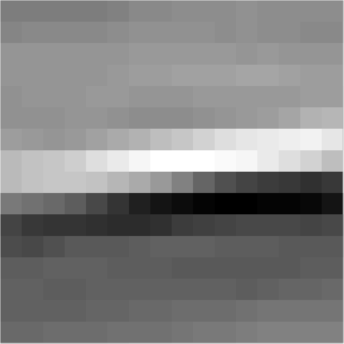}}
\subfloat{\makebox[0.9em]{\small $...$}}
\subfloat[0.122]{\vcenteredinclude{width=0.115\columnwidth}{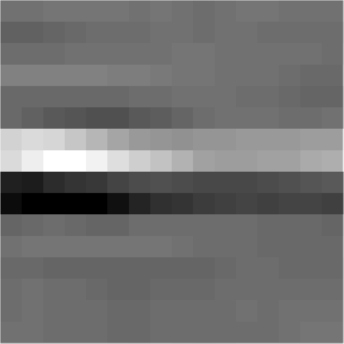}}
\subfloat{\makebox[0.9em]{\small $...$}}
\subfloat[0.083]{\vcenteredinclude{width=0.115\columnwidth}{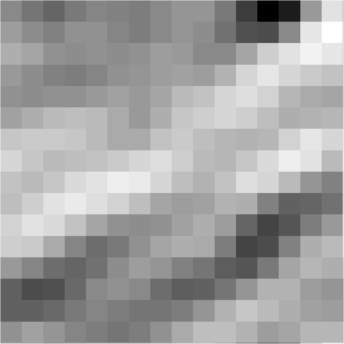}}
\subfloat{\makebox[0.9em]{\small $...$}}
\subfloat[0.039]{\vcenteredinclude{width=0.115\columnwidth}{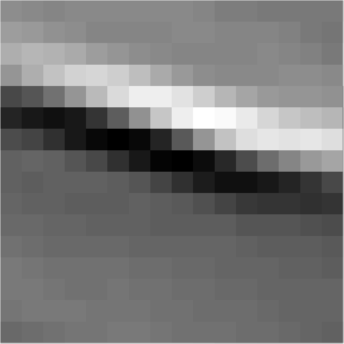}}
\subfloat{\makebox[0.9em]{\small $...$}}
\subfloat[0.024]{\vcenteredinclude{width=0.115\columnwidth}{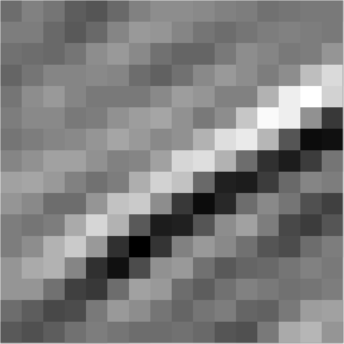}}
\subfloat{\makebox[0.9em]{\small $...$}}
\subfloat[0.001]{\vcenteredinclude{width=0.115\columnwidth}{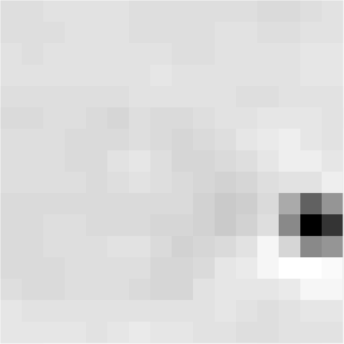}}

\vspace{-0.8em}
\hspace{0.3em}
\subfloat{\makebox[2.5em]{\footnotesize (d)}}
\subfloat[$\hat{\boldsymbol{x}}$]{\vcenteredinclude{width=0.115\columnwidth}{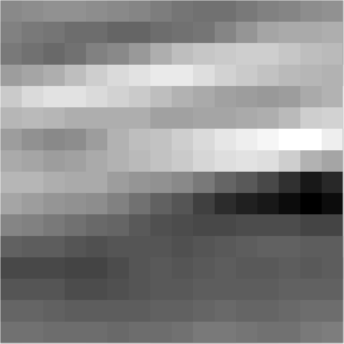}}
\subfloat{\makebox[0.9em]{\small =}}
\subfloat[0.194]{\vcenteredinclude{width=0.115\columnwidth}{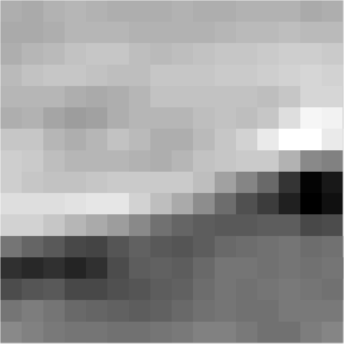}}
\subfloat{\makebox[0.9em]{\small +}}
\subfloat[0.164]{\vcenteredinclude{width=0.115\columnwidth}{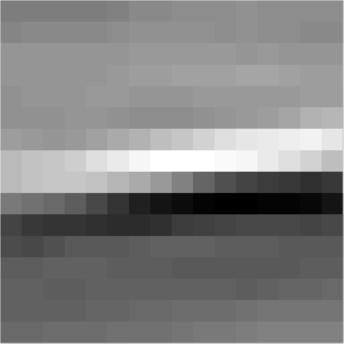}}
\subfloat{\makebox[0.9em]{\small +}}
\subfloat[0.141]{\vcenteredinclude{width=0.115\columnwidth}{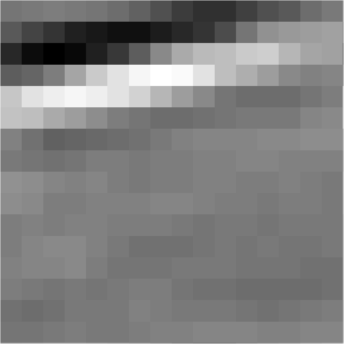}}
\subfloat{\makebox[0.9em]{\small +}}
\subfloat[0.066]{\vcenteredinclude{width=0.115\columnwidth}{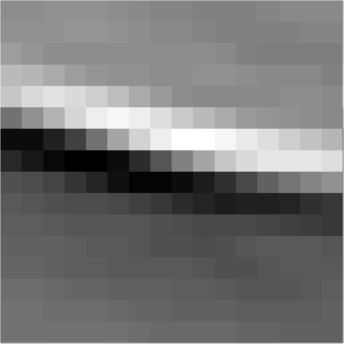}}
\subfloat{\makebox[0.9em]{\small +}}
\subfloat[0.026]{\vcenteredinclude{width=0.115\columnwidth}{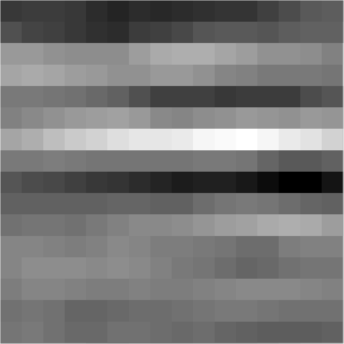}}
\subfloat{\makebox[0.9em]{\small +}}
\subfloat[0.009]{\vcenteredinclude{width=0.115\columnwidth}{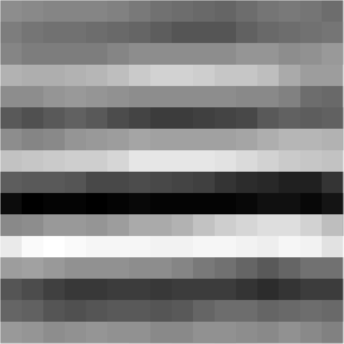}}
\end{minipage}
\caption{An example of the ``explaining away'' conditional dependence
provided by optimization-based inference. Sparse 
representations constructed by feed-forward nonnegative soft thresholding 
(a) have many more non-zero elements due to redundancy and spurious 
activations (c). On the other hand, sparse representations found by $\ell_1$-penalized, 
nonnegative least-squares optimization (b) yield a more parsimonious set of components (d) that 
optimally reconstruct approximations of the data.  
}
\label{fig:explaining_away}
\end{figure}

Deep convolutional neural networks have achieved remarkable success in the field of
computer vision. While far from new~\cite{lecun1998gradient}, the increasing availability of 
extremely large, labeled datasets along with modern advances in computation with
specialized hardware have resulted in
state-of-the-art performance in many problems, including essentially all
visual learning tasks. Examples include image classification~\cite{huang2017densely}, 
object detection~\cite{huang2017speed}, and semantic segmentation~\cite{chen2017deeplab}. 
Despite a rich history of practical and theoretical insights about these problems, modern deep 
learning techniques typically rely on task-agnostic models and poorly-understood heuristics.
However, recent work~\cite{lin2016inverse, tulsiani2017multi, brachmann2016dsac} has 
shown that specialized architectures incorporating classical domain 
knowledge can increase parameter efficiency, relax training data requirements, and improve 
performance.

Prior to the advent of modern deep learning, optimization-based methods like 
component analysis and sparse coding dominated the field of representation learning. 
These techniques use structured matrix factorization to decompose 
data into linear combinations of shared components. Latent representations 
are inferred by minimizing reconstruction error subject to constraints that enforce 
properties like uniqueness and interpretability.
Unlike feed-forward alternatives that construct 
representations in closed-form via independent feature detectors, this 
optimization-based approach 
naturally introduces conditional dependence between features in order to  
best explain data, a useful phenomenon commonly referred to as 
``explaining away'' within the context of graphical models~\cite{bengio2013representation}.
An example of this effect is shown in Fig.~\ref{fig:explaining_away}, which compares sparse 
representations constructed using feed-forward soft thresholding with those given by  
optimization-based inference with an $\ell_1$ penalty. While many components in an overcomplete 
set of features may have high-correlation with an image, constrained   
optimization introduces competition between components resulting in more parsimonious representations. 

Component analysis methods are also often guided by intuitive goals of incorporating 
prior knowledge into learned representations. For example, statistical 
independence allows for the separation of signals into distinct generative 
sources~\cite{jutten1991blind}, non-negativity leads to  parts-based decompositions of
objects~\cite{lee1999learning}, and sparsity gives rise to locality and frequency 
selectivity~\cite{olshausen1996emergence}. Due to the difficulty of enforcing intuitive 
constraints like these with feed-forward computations, deep learning architectures are instead 
often motivated by distantly-related biological systems~\cite{simonyan2014two} or  
poorly-understand internal mechanisms such as covariate shift~\cite{ioffe2015batch} 
and gradient flow~\cite{he2016identity}. 
Furthermore, while a theoretical understanding 
of deep learning is fundamentally lacking~\cite{zhang2017understanding}, even non-convex
formulations of matrix factorization are often associated with guarantees of 
convergence~\cite{bao2016dictionary}, generalization~\cite{liu2016dimensionality}, 
uniqueness~\cite{gillis2012sparse}, and even global optimality~\cite{haeffele2014structured}.

In order to unify the intuitive and theoretical insights of component analysis with the 
practical advances made possible through deep learning, we introduce the framework of 
Deep Component Analysis (DeepCA). This novel model formulation can be interpreted as a
multilayer extension of traditional component analysis in which multiple layers are
learned jointly with intuitive constraints intended to encode structure and prior knowledge.
DeepCA can also be motivated from the perspective of deep neural networks by relaxing the 
implicit assumption that the input to a layer is constrained to be the output of 
the previous layer, as shown in Eq.~\ref{eq:decouple} below.
In a feed-forward network (left), the output of layer $j$, denoted $\boldsymbol{a}_j$, is 
given in closed-form as a nonlinear function of $\boldsymbol{a}_{j-1}$.
DeepCA (right) instead takes a generative approach in which the latent variables 
 $\boldsymbol{w}_j$ 
associated with layer $j$ are \emph{inferred} to optimally reconstruct $\boldsymbol{w}_{j-1}$ 
as a linear combination of learned components subject to some constraints $\mathcal{C}_j$.
\begin{equation}
\text{Feed-Forward: }\boldsymbol{a}_{j}=\phi(\mathbf{B}_{j}^{\mathsf{T}}\boldsymbol{a}_{j-1}) 
\, \Longrightarrow \, \text{DeepCA: }
\mathbf{B}_{j}\boldsymbol{w}_{j} \approx \boldsymbol{w}_{j-1}\text{ s.t. }\boldsymbol{w}_j \in \mathcal{C}_j
\label{eq:decouple}
\end{equation}

\begin{figure}[tb]
\centering
\hspace{\fill}
\subfloat[Feed-Forward]{
\includegraphics[height=4.2cm]{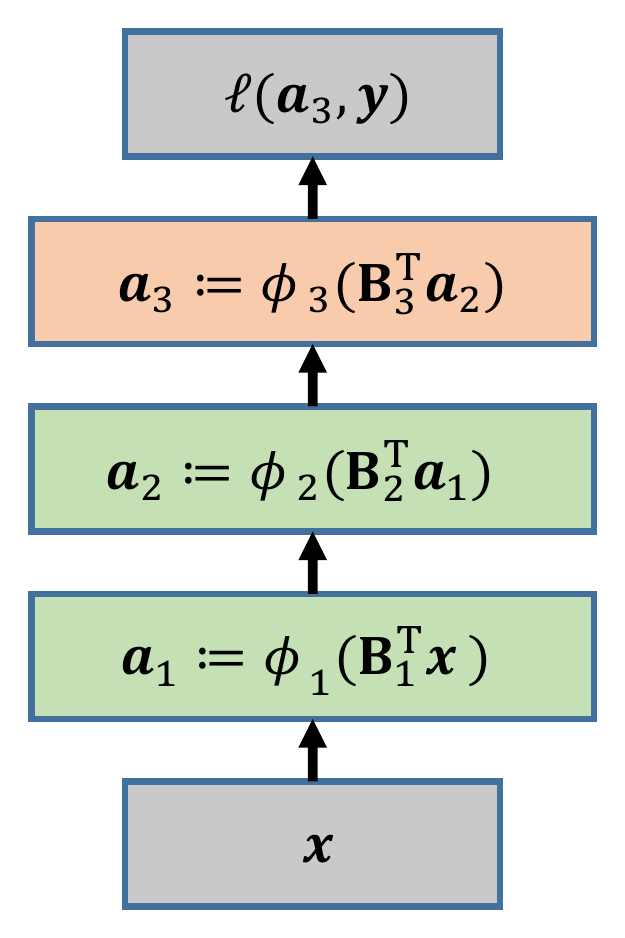}
}
\hspace{\fill}
\subfloat[DeepCA]{
\includegraphics[height=4.2cm]{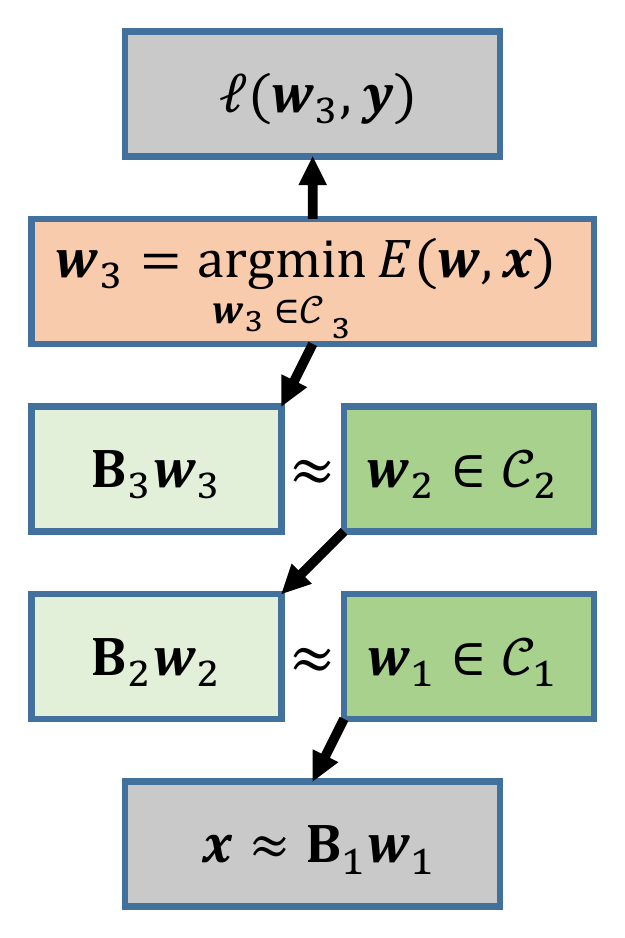}
}
\hspace{\fill}
\subfloat[ADNN]{
\includegraphics[height=4.2cm]{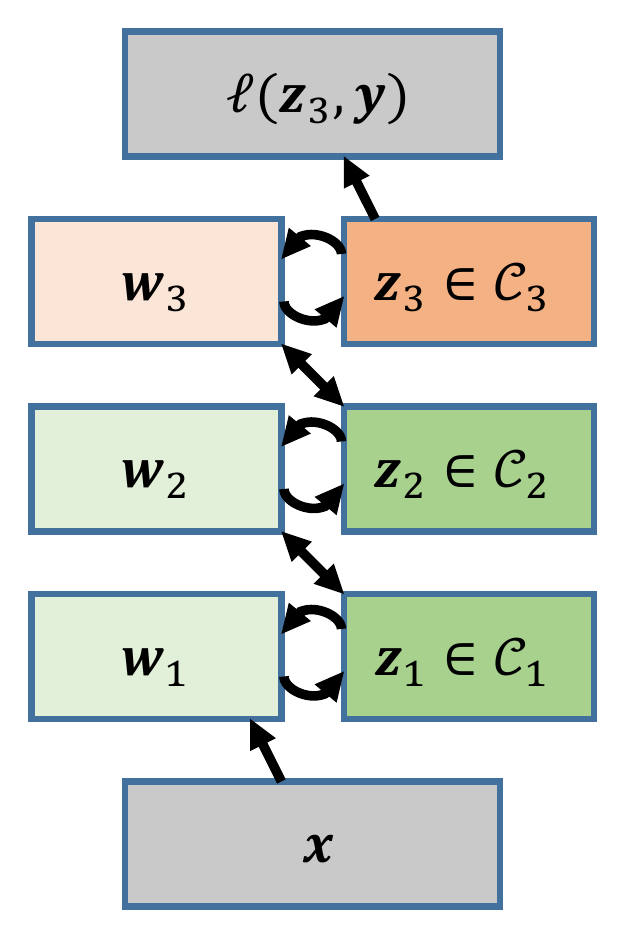}
}
\hspace{\fill}
\subfloat[Unrolled ADNN]{
\includegraphics[height=4.2cm]{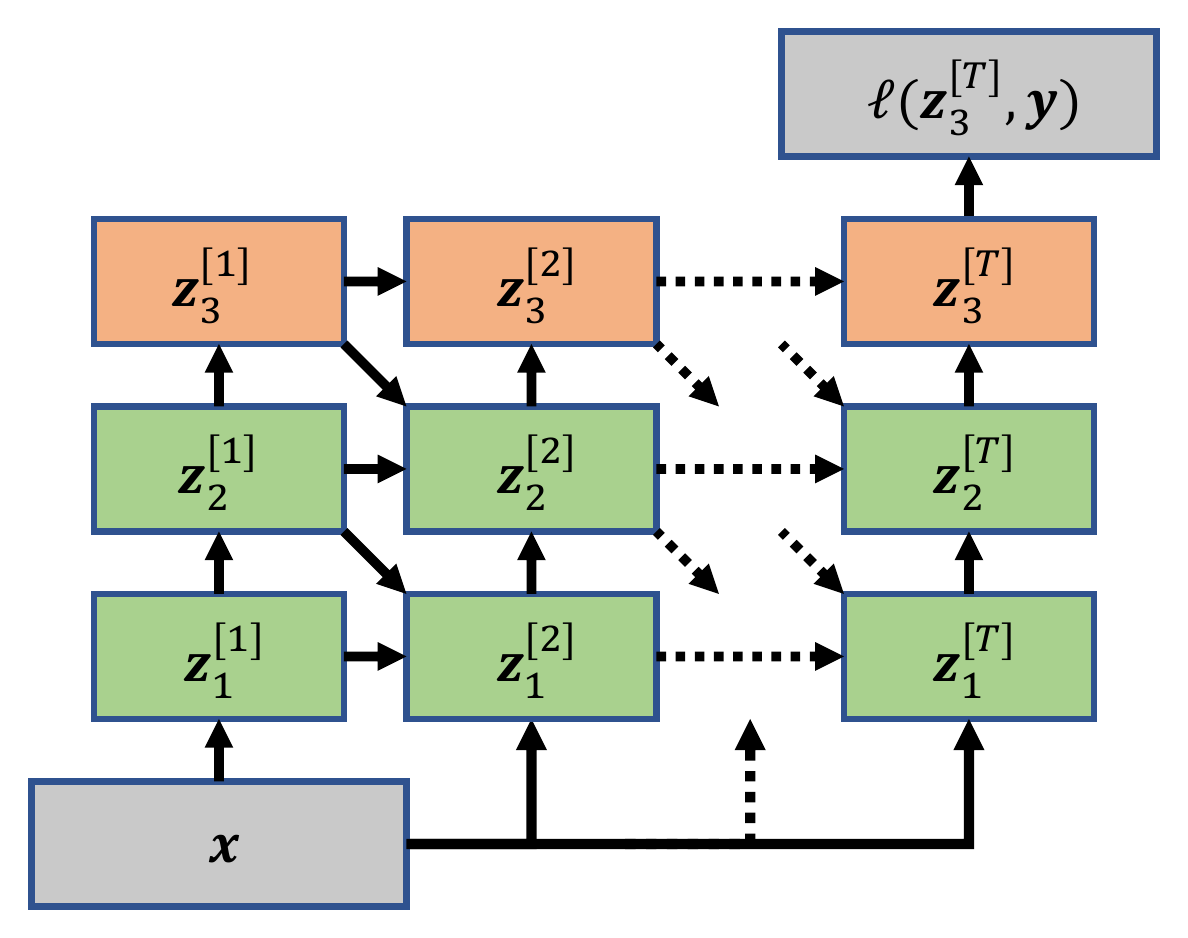}
}
\hspace{\fill}
\caption{
A comparison between feed-forward neural networks and the proposed
deep component analysis (DeepCA) model. While standard deep networks construct
learned representations as feed-forward compositions of nonlinear functions (a), 
DeepCA instead treats them as unknown latent variables to be inferred by 
constrained optimization (b). To accomplish this, we 
propose a differentiable inference algorithm that can be expressed as an 
Alternating Direction Neural Network (ADNN), a recurrent generalization of 
feed-forward networks (c) that can be unrolled to a fixed number of iterations for 
learning via backpropagation (d).
}
\label{fig:front}
\end{figure}

From this perspective, intermediate network ``activations'' cannot be found in closed-form 
but instead require explicitly solving an optimization problem. 
While a variety of different techniques could be used for performing this inference, we propose
the Alternating Direction Method of Multipliers (ADMM)~\cite{boyd2011distributed}. 
Importantly, we demonstrate that 
after proper initialization, a single iteration of this algorithm is equivalent to a
pass through an associated feed-forward neural network with nonlinear activation functions
interpreted as proximal operators corresponding to penalties or constraints on the 
coefficients. The full inference procedure can thus be implemented using 
Alternating Direction Neural Networks (ADNN), recurrent generalizations of feed-forward 
networks that allow for parameter learning using backpropagation. A comparison 
between standard neural networks and DeepCA is shown in Fig.~\ref{fig:front}.
Experimentally, 
we demonstrate that recurrent passes through convolutional neural networks
enable better sparsity control resulting in consistent performance improvements in both 
supervised and unsupervised tasks without introducing any additional parameters.

\begin{figure}[tb]
\centering

\captionsetup[subfigure]{labelformat=empty,justification=centering}
\subfloat[]{
\includegraphics[width=0.104\textwidth]{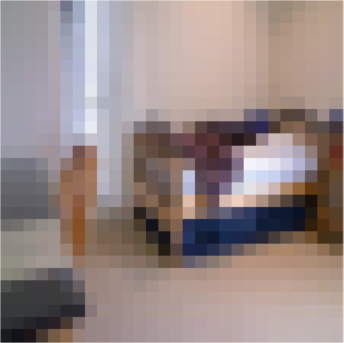}}
\subfloat[]{
\includegraphics[width=0.104\textwidth]{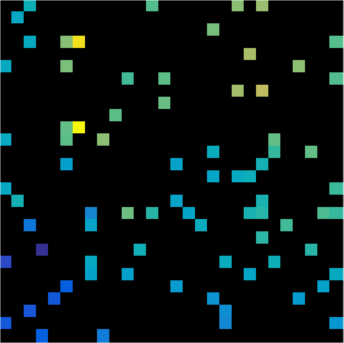}} 
\subfloat[]{
\includegraphics[width=0.104\textwidth]{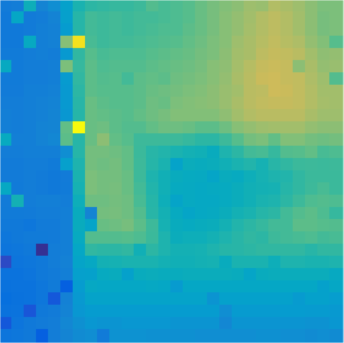}} 
\subfloat[]{
\includegraphics[width=0.104\textwidth]{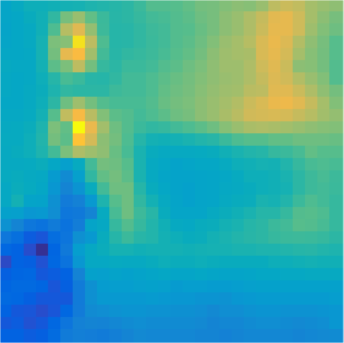}} 
\subfloat[]{
\includegraphics[width=0.104\textwidth]{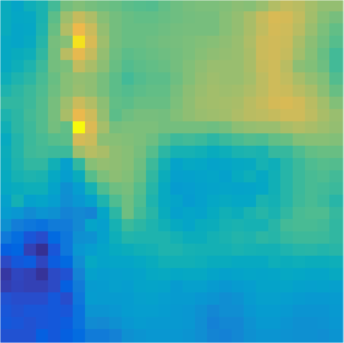}} 
\subfloat[]{
\includegraphics[width=0.104\textwidth]{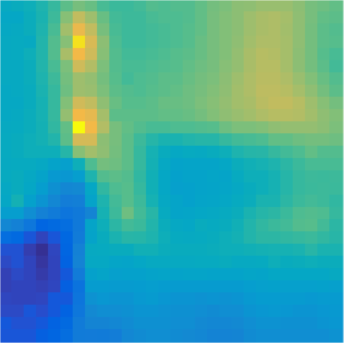}} 
\subfloat[]{
\includegraphics[width=0.104\textwidth]{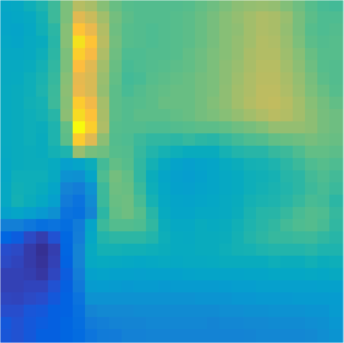}} 
\subfloat[]{
\includegraphics[width=0.104\textwidth]{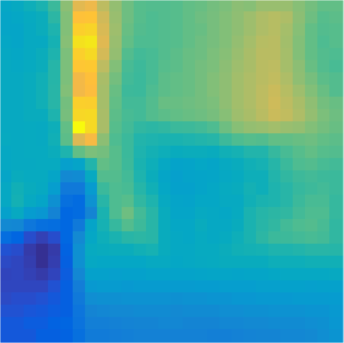}} 
\subfloat[]{
\includegraphics[width=0.104\textwidth]{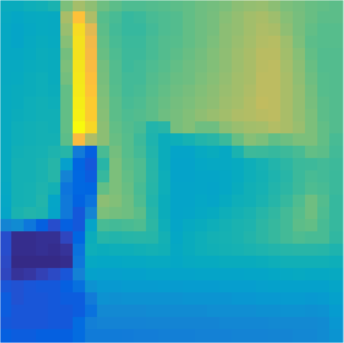}} 
\vspace{-2.15em}
\subfloat[]{
\includegraphics[width=0.104\textwidth]{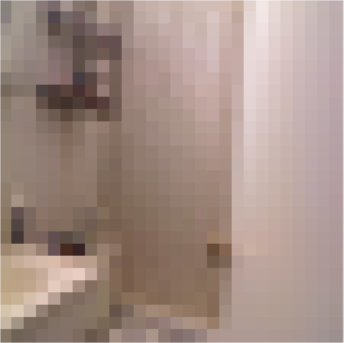}}
\subfloat[]{
\includegraphics[width=0.104\textwidth]{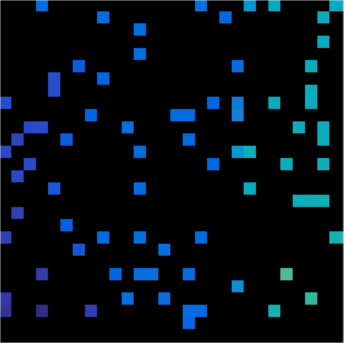}} 
\subfloat[]{
\includegraphics[width=0.104\textwidth]{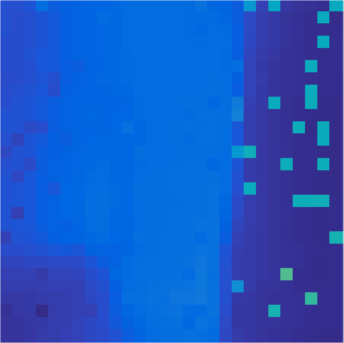}} 
\subfloat[]{
\includegraphics[width=0.104\textwidth]{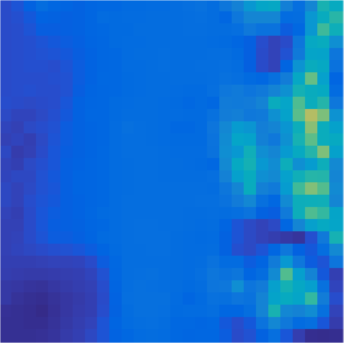}} 
\subfloat[]{
\includegraphics[width=0.104\textwidth]{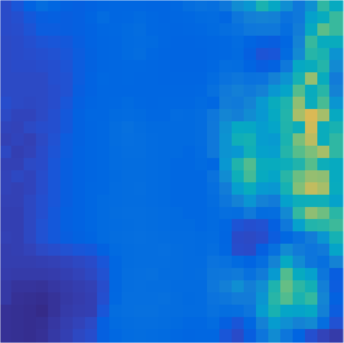}} 
\subfloat[]{
\includegraphics[width=0.104\textwidth]{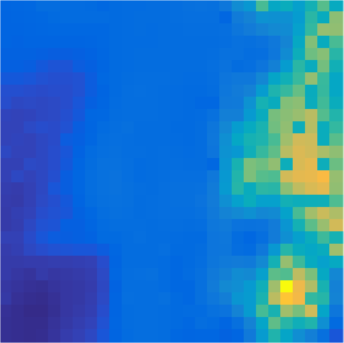}} 
\subfloat[]{
\includegraphics[width=0.104\textwidth]{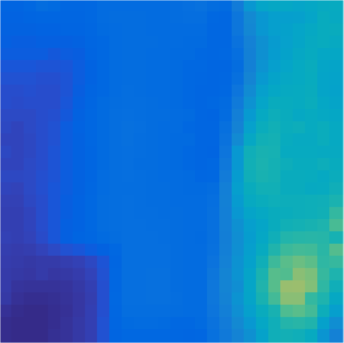}} 
\subfloat[]{
\includegraphics[width=0.104\textwidth]{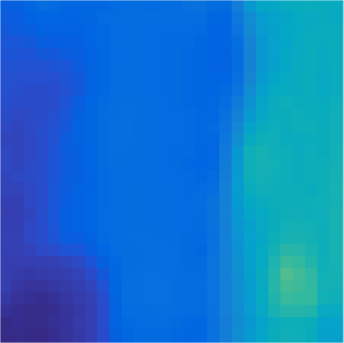}} 
\subfloat[]{
\includegraphics[width=0.104\textwidth]{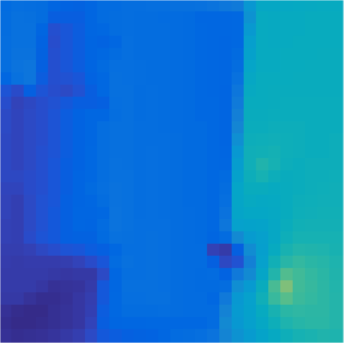}} 
\vspace{-2.15em}
\subfloat[]{
\includegraphics[width=0.104\textwidth]{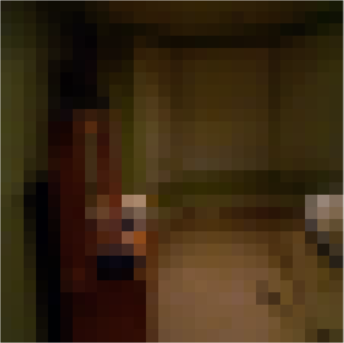}}
\subfloat[]{
\includegraphics[width=0.104\textwidth]{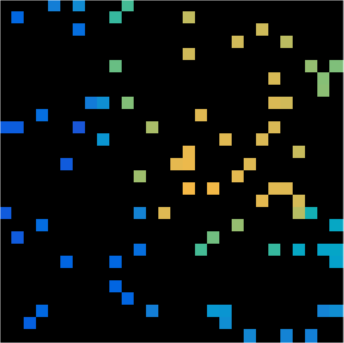}} 
\subfloat[]{
\includegraphics[width=0.104\textwidth]{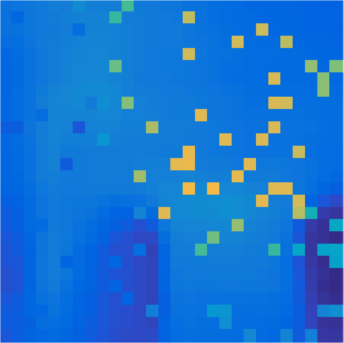}} 
\subfloat[]{
\includegraphics[width=0.104\textwidth]{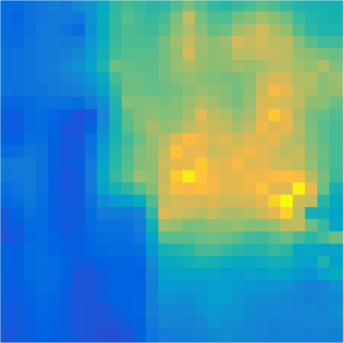}} 
\subfloat[]{
\includegraphics[width=0.104\textwidth]{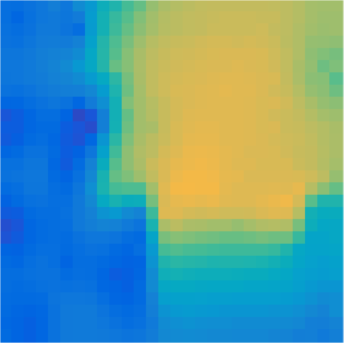}} 
\subfloat[]{
\includegraphics[width=0.104\textwidth]{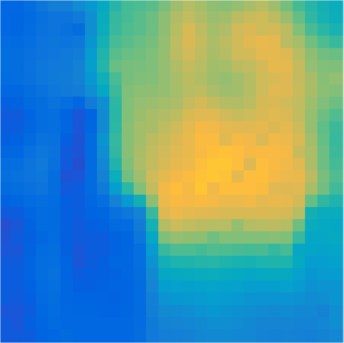}} 
\subfloat[]{
\includegraphics[width=0.104\textwidth]{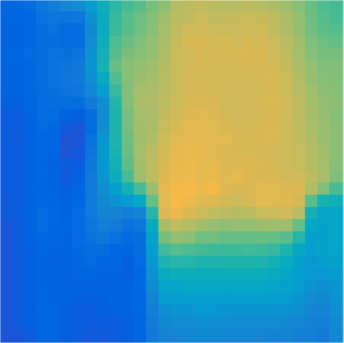}} 
\subfloat[]{
\includegraphics[width=0.104\textwidth]{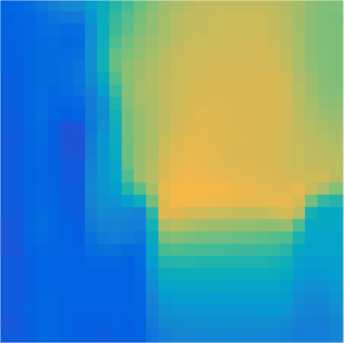}} 
\subfloat[]{
\includegraphics[width=0.104\textwidth]{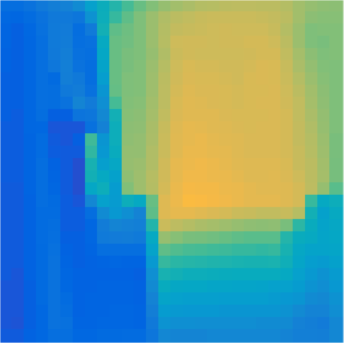}} 
\vspace{-2.15em}
\subfloat[ Image]{
\includegraphics[width=0.104\textwidth]{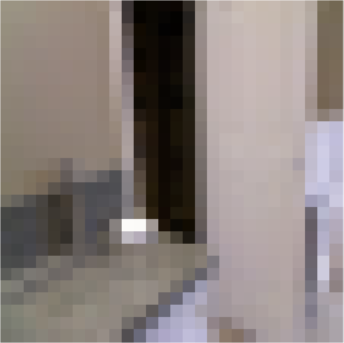}}
\subfloat[ Given]{
\includegraphics[width=0.104\textwidth]{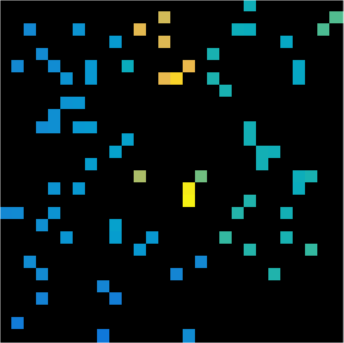}} 
\subfloat[ Baseline]{
\includegraphics[width=0.104\textwidth]{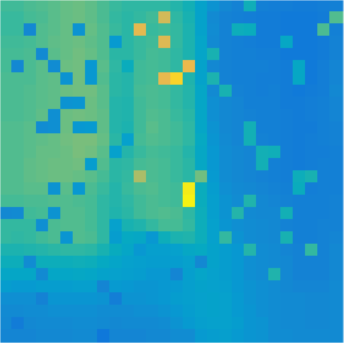}} 
\subfloat[ $\boldsymbol{T=2}$]{
\includegraphics[width=0.104\textwidth]{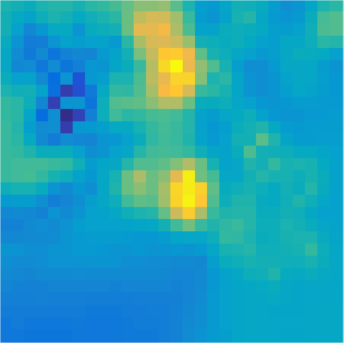}} 
\subfloat[ $\boldsymbol{T=3}$]{
\includegraphics[width=0.104\textwidth]{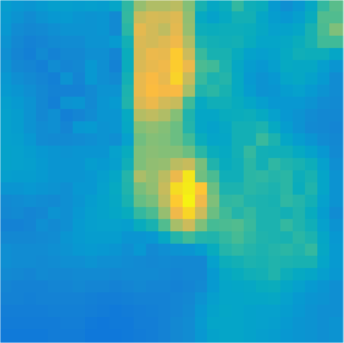}} 
\subfloat[ $\boldsymbol{T=5}$]{
\includegraphics[width=0.104\textwidth]{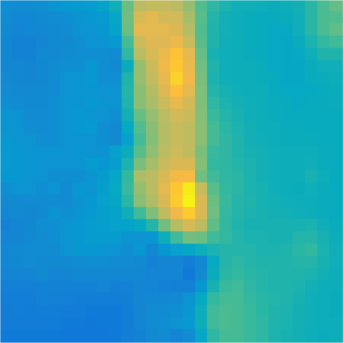}} 
\subfloat[ $\boldsymbol{T=10}$]{
\includegraphics[width=0.104\textwidth]{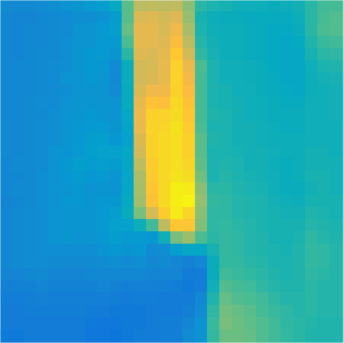}} 
\subfloat[ $\boldsymbol{T=20}$]{
\includegraphics[width=0.104\textwidth]{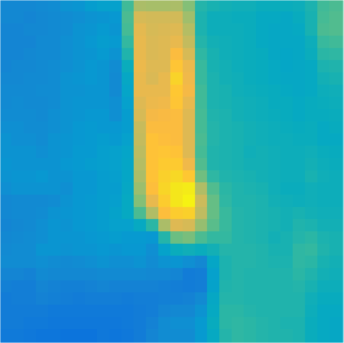}} 
\subfloat[ Truth]{
\includegraphics[width=0.104\textwidth]{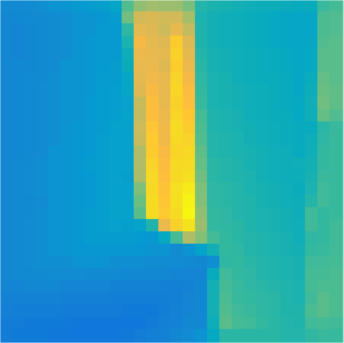}} 

\caption{
A demonstration of DeepCA applied to single-image depth prediction using
images concatenated with sparse sets of known depth values as input. 
Baseline feed-forward networks are not guaranteed to produce outputs that are 
consistent with the given depth values. ADNNs with an
increasing number of 
iterations ($T>1$) learn to satisfy the sparse output constraints, resolving 
ambiguities for more accurate predictions without unrealistic discontinuities. 
}
\label{fig:iter_viz}
\end{figure}

More importantly, DeepCA also allows for other 
constraints that would be impossible to effectively enforce with a single feed-forward 
pass through a network. As an example, we consider the task of single-image depth prediction, a
difficult problem due to the absence of three-dimensional information
such as scale and perspective. In many 
practical scenarios, however, sparse sets of known depth outputs are available for resolving 
these ambiguities to improve accuracy. This prior knowledge can come from additional sensor 
modalities like LIDAR or from other 3D reconstruction algorithms that provide sparse 
depths around textured image regions. Feed-forward networks have been 
proposed for this problem by concatenating known depth values as an additional input 
channel~\cite{ma2017sparse}. However, while this provides useful context, 
predictions are not guaranteed to be consistent with the given outputs leading to 
unrealistic discontinuities. In comparison, DeepCA enforces the 
constraints by treating predictions as unknown latent 
variables. Some examples of how this behavior can resolve ambiguities are shown in 
Fig.~\ref{fig:iter_viz} where ADNNs with additional iterations learn to propagate information 
from the given depth values to produce more accurate predictions.

In addition to practical advantages, our model also provides a novel perspective for 
conceptualizing deep learning techniques. Due to the decoupling of layers
provided by relaxing the feed-forward function composition constraints, DeepCA can
be equivalently expressed as a shallow model augmented with architecture-dependent 
structure imposed on the model parameters. In the case of rectified linear 
unit (ReLU) activation functions~\cite{glorot2011deep}, this allows for the direct 
application of results from sparse approximation theory, suggesting new insights towards better 
understanding why deep neural networks are so effective. 

\section{Background and Related Work}

In order to motivate our approach, we first provide some
background on matrix factorization, component analysis, and 
deep neural networks.

\subsection{Component Analysis and Matrix Factorization}

Component analysis is a common approach for shallow representation learning that 
approximately decomposes data $\boldsymbol{x}\in\mathbb{R}^d$ into linear combinations of 
learned components in $\mathbf{B}\in\mathbb{R}^{d\times k}$. This is typically accomplished
by minimizing reconstruction error subject to
constraints $\mathcal{C}$  on the coefficients that serve to 
resolve ambiguity or incorporate prior knowledge such as low-rank
structure or sparsity.
Some examples include Principal Component Analysis (PCA)~\cite{wold1987principal} for
dimensionality reduction and sparse dictionary 
learning~\cite{bao2016dictionary} which accommodates overcomplete representations by enforcing sparsity. 

While component analysis problems are 
typically non-convex, their structure naturally suggests simple
alternating minimization strategies that are often guaranteed to converge~\cite{xu2013block}.
However, unlike backpropagation with stochastic gradient descent, these techniques 
typically require careful initialization in order to avoid poor local minima. 
Alternatively, we consider a nested optimization problem that separates learning from inference: 
\begin{equation}
\underset{\mathbf{B}}{\arg\min}\sum_{i=1}^{n}\tfrac{1}{2}\lVert \boldsymbol{x}^{(i)}-\mathbf{B}\boldsymbol{f}(\boldsymbol{x}^{(i)})\rVert _{2}^{2}\quad\text{s.t. }\boldsymbol{f}(\boldsymbol{x})=\underset{\boldsymbol{w}\in \mathcal{C}}{\arg\min}\,\tfrac{1}{2}\left\lVert \boldsymbol{x}-\mathbf{B}\boldsymbol{w}\right\rVert _{2}^{2}
\label{eq:ca_inference}
\end{equation}
Here, the inference function 
$\boldsymbol{f}:\mathbb{R}^d\rightarrow\mathbb{R}^k$ is a potentially nonlinear 
transformation that maps data to their corresponding representations by solving an optimization 
problem with fixed parameters.
For unconstrained PCA with orthogonal components, this inference problem 
has a simple closed-form solution given by the linear transformation 
$\boldsymbol{f}^{\mathrm{PCA}}(\boldsymbol{x})=\mathbf{B}^\mathsf{T}\boldsymbol{x}$.
Substituting this into Eq.~\ref{eq:ca_inference} results in a linear autoencoder with one 
hidden layer and tied weights, which has the same unique global minimum but can be trained 
by backpropagation~\cite{baldi1989neural}. 

With general constraints, inference typically cannot be accomplished in closed form but must 
instead rely on an iterative optimization algorithm. However, if this algorithm is 
composed as a finite sequence of differentiable transformations, then the model parameters can 
still be learned in the same way by backpropagating gradients through the steps of the inference 
algorithm. We extend this idea by representing an algorithm for inference in our DeepCA model 
as a recurrent neural network unrolled to a fixed number of iterations.

\subsection{Deep Neural Networks}

Recently, deep neural networks have emerged as the preferred alternative to component 
analysis for representation learning of visual data. Their ability to jointly learn multiple 
layers of abstraction has been shown to allow for encoding increasingly complex features such as 
textures and object parts~\cite{lee2009convolutional}. Unlike with component analysis, inference is
given in closed-form by design. Specifically, a representation is constructed
by passing an image $\boldsymbol{x}$ through the composition of alternating linear 
transformations with parameters $\mathbf{B}_{j}$ and $\boldsymbol{b}_{j}$ and 
fixed nonlinear activation functions $\phi_j$ for layers $j=1,\dotsc,l$ as follows: 
\begin{equation}
\boldsymbol{f}^{\mathrm{DNN}}(\boldsymbol{x})=\phi_l\big(\mathbf{B}_{l}^{\mathsf{T}}\cdots\phi_2(\mathbf{B}_{2}^{\mathsf{T}}(\phi_1(\mathbf{B}_{1}^{\mathsf{T}}\boldsymbol{x}-\boldsymbol{b}_{1})-\boldsymbol{b}_{2})\cdots-\boldsymbol{b}_{l}\big)
\label{eq:feed_forward}
\end{equation}


Instead of considering the forward pass of a neural network
as an arbitrary nonlinear function, we interpret it as a method for approximate 
inference in an unsupervised generative model. This follows from previous work
which has shown it to be equivalent to bottom-up inference in a probabilistic graphical
model~\cite{patel2016probabilistic} or   
approximate inference 
in a multi-layer convolutional sparse coding model~\cite{papyan2017convolutional, sulam2017pursuit}.
However, these approaches have limited practical applicability due to their reliance on careful 
hyperparameter selection and specialized optimization algorithms. While ADMM has
been proposed as a gradient-free alternative to backpropagation for parameter 
learning~\cite{taylor2016training}, we use it only for inference which allows for simpler learning
using backpropagation with arbitrary loss functions.

Aside from ADNNs, recurrent feedback has been proposed in other models to improve performance 
by iteratively refining predictions, especially for applications such as human pose estimation
or image segmentation where outputs have complex correlation 
patterns~\cite{carreira2016human, belagiannis2017recurrent, li2016iterative}.
While some methods also implement feedback by directly unrolling iterative algorithms, 
they are often geared towards specific applications such as graphical model 
inference~\cite{chen2015learning, hu2016bottom}, solving under-determined inverse 
problems~\cite{gregor2010fastsparse, diamond2017unrolled, sun2016deep}, or  
image alignment~\cite{lin2016inverse}. Similar to~\cite{zamir2017feedback},
DeepCA provides a general mechanism for feedback in arbitrary 
neural networks, but it is motivated by the more interpretable goal of minimizing reconstruction
error subject to constraints on network activations.

\section{Deep Component Analysis}


Deep Component Analysis generalizes the shallow inference objective 
function in Eq.~\ref{eq:ca_inference} by introducing additional layers $j=1,\dots,l$ with
parameters $\mathbf{B}_j\in\mathbb{R}^{p_{j-1}\times p_j}$. 
Optimal DeepCA inference is then accomplished by solving:
\begin{equation}
\boldsymbol{f}^{*}(\boldsymbol{x})=\underset{\{\boldsymbol{w}_{j}\}}{\arg\min}\sum_{j=1}^{l}\tfrac{1}{2}\left\Vert \boldsymbol{w}_{j-1}-\mathbf{B}_{j}\boldsymbol{w}_{j}\right\Vert _{2}^{2}+\Phi_{j}(\boldsymbol{w}_{j})\quad\text{s.t. }\boldsymbol{w}_0=\boldsymbol{x}
\label{eq:deepca}
\end{equation}
Instead of constraint sets $\mathcal{C}_j$, we use penalty functions 
$\Phi_{j}:\mathbb{R}^{p_j}\rightarrow\mathbb{R}$ to enable more general priors. Note that hard constraints can still be represented by indicator functions 
$I(\boldsymbol{w}_j \in \mathcal {C}_j)$ that equal zero if 
$\boldsymbol{w}_j \in \mathcal {C}_j$ and infinity otherwise. 
While we use pre-multiplication with a weight matrix $\mathbf{B}_{j}$ to simplify notation, our 
method also supports any linear transformation by replacing 
transposed weight matrix multiplication with its corresponding adjoint operator. For example, 
the adjoint of convolution is transposed convolution, a popular approach to upsampling in 
convolutional networks~\cite{noh2015learning}. 

If the penalty functions are convex, this problem is also convex and can be
solved using standard optimization methods. While this appears to differ
substantially from inference in deep neural networks, we later show that it can be seen as a 
generalization of the feed-forward inference function in Eq.~\ref{eq:feed_forward}.
In the remainder of this section, we 
justify the use of penalty functions in lieu of explicit nonlinear activation functions 
by drawing connections between non-negative $\ell_1$ regularization and ReLU activation 
functions. We then propose a general algorithm for solving Eq.~\ref{eq:deepca} for
the unknown coefficients and formalize the relationship between DeepCA and 
traditional deep neural networks, which enables parameter learning via backpropagation.

\subsection{From Activation Functions to Constraints} \label{sec:constraints}

Before introducing our inference algorithm, we first discuss the connection between penalties and their 
nonlinear proximal operators, which forms the basis of the close relationship between DeepCA and 
traditional neural networks. Ubiquitous within the field of convex optimization, 
proximal algorithms~\cite{parikh2014proximal} are methods for solving nonsmooth optimization problems. 
Essentially, these techniques work by breaking a problem down 
into a sequence of smaller problems that can often be solved in closed-form 
by proximal operators
$\mathrm{\phi}:\mathbb{R}^d\rightarrow\mathbb{R}^d$ associated with penalty functions 
$\Phi:\mathbb{R}^d\rightarrow\mathbb{R}$ given by the solution to the following optimization problem, which generalizes projection onto a constraint set: 
\begin{equation}
\mathrm{\phi}(\boldsymbol{w})=\underset{\boldsymbol{w}^{\prime}}{\arg\min}\,\tfrac{1}{2}\left\lVert \boldsymbol{w}-\boldsymbol{w}^{\prime}\right\rVert _{2}^{2}+\Phi(\boldsymbol{w}^{\prime})
\label{eq:prox}
\end{equation}

Within the framework of DeepCA, we interpret nonlinear activation functions in deep 
networks as proximal operators associated with convex penalties on latent coefficients in each layer. While this connection 
cannot be used to generalize all nonlinearities, many can naturally be 
interpreted as proximal operators. For example, the sparsemax activation function is a  
projection onto the probability simplex~\cite{martins2016softmax}. Similarly, the 
ReLU activation function is a projection onto the nonnegative orthant. When used  
with a negative bias $\boldsymbol{b}$, it is equivalent to nonnegative soft-thresholding
$\mathcal{S}^+_{\boldsymbol{b}}$, the proximal operator associated with nonnegative 
$\mathcal{\ell}_1$ regularization:
\begin{equation}
\Phi^{\ell_{1}^{+}}(\boldsymbol{w})=I(\boldsymbol{w}\geq0)+{\textstyle \sum}_{p}b_{p}\left|w_{p}\right| \quad \Longrightarrow \quad
\mathrm{\phi}^{\ell_{1}^{+}}(\boldsymbol{w})=\mathcal{S}^+_{\boldsymbol{b}}(\boldsymbol{w})=\mathrm{ReLU}(\boldsymbol{w}-\boldsymbol{b})
\label{eq:prox_relu}
\end{equation}
While this equivalence has been noted previously as a means to theoretically analyze 
convolutional neural networks~\cite{papyan2017convolutional}, DeepCA supports optimizing
the bias $\boldsymbol{b}$ as an $\ell_1$ penalty hyperparameter via backpropagation 
for adaptive regularization, which results in better control of representation sparsity.

In addition to standard activation 
functions, DeepCA also allows for enforcing additional constraints that encode prior knowledge. 
For our example of single-image depth prediction with a sparse set of known 
outputs $\boldsymbol{y}$ provided as prior knowledge, the penalty function on the final output 
$\boldsymbol{w}_l$ is 
$\boldsymbol{\Phi}_l(\boldsymbol{w}_l) = I(\mathbf{S}\boldsymbol{w}_l = \boldsymbol{y})$ 
where the selector matrix $\mathbf{S}$ extracts the indices corresponding to the known outputs 
in $\boldsymbol{y}$. The associated proximal operator $\boldsymbol{\phi}_l$ projects onto this 
constraint set by simply correcting the outputs that disagree with the known constraints. 
Note that this would not be an effective output nonlinearity in a feed-forward network because, 
while the constraints would be technically satisfied, there is nothing to enforce that they be 
consistent with neighboring predictions leading to unrealistic discontinuities. In contrast, 
DeepCA inference minimizes the reconstruction error at each layer subject to these constraints
by taking multiple iterations through the network.

\subsection{Inference by the Alternating Direction Method of Multipliers}

With the model parameters fixed, we solve our DeepCA inference problem using the Alternating 
Direction Method of Multipliers (ADMM), a general optimization technique that has 
been successfully used in a wide variety of applications~\cite{boyd2011distributed}. 
To derive the 
algorithm applied to our problem, we first modify our objective function by introducing 
auxiliary variables $\boldsymbol{z}_j$ that we constrain to be equal to the unknown coefficients 
$\boldsymbol{w}_j$, as shown in Eq.~\ref{eq:deepca_z} below.
\begin{equation}
\underset{\{\boldsymbol{w}_{j},\boldsymbol{z}_{j}\}}{\arg\min}\sum_{j=1}^{l}\tfrac{1}{2}\left\Vert \boldsymbol{z}_{j-1}{-}\mathbf{B}_{j}\boldsymbol{w}_{j}\right\Vert _{2}^{2}+\Phi_{j}(\boldsymbol{z}_{j})\quad\text{s.t. }\boldsymbol{w}_{0}=\boldsymbol{x},\,\forall j:\boldsymbol{w}_{j}=\boldsymbol{z}_{j}
\label{eq:deepca_z}
\end{equation}

From this, we construct the augmented Lagrangian $\mathcal{L}_\rho$ with dual variables 
$\boldsymbol{\lambda}$ and a quadratic penalty hyperparameter $\rho=1$ that can affect 
convergence speed: 
\begin{equation}
\mathcal{L}_\rho=\sum_{j=1}^{l}\tfrac{1}{2}\left\Vert \boldsymbol{z}_{j-1}-\mathbf{B}_{j}\boldsymbol{w}_{j}\right\Vert _{2}^{2}+\Phi_{j}(\boldsymbol{z}_{j})
+\boldsymbol{\lambda}_{j}^{\mathsf{T}}(\boldsymbol{w}_{j}-\boldsymbol{z}_{j})+\tfrac{\rho}{2}\left\lVert \boldsymbol{w}_{j}-\boldsymbol{z}_{j}\right\rVert _{2}^{2}
\label{eq:lagrangian}
\end{equation}

The ADMM algorithm then proceeds by iteratively minimizing $\mathcal{L}_\rho$ with respect to 
each set of variables with the others fixed, breaking our full inference optimization problem into smaller
pieces that can each be solved in closed form. Due to the decoupling of layers in 
our DeepCA model, the latent activations can be updated incrementally by stepping  
through each layer in succession, resulting in faster convergence 
and updates that mirror the computational structure of deep neural networks. With only one layer,
our objective function is separable and so this algorithm reduces to the classical two-block
ADMM, which has extensive convergence guarantees~\cite{boyd2011distributed}. For multiple 
layers, however, this algorithm can be seen as  
an instance of the cyclical multi-block ADMM with quadratic coupling terms. While our 
experiments have shown this approach to be effective in our applications, theoretical 
analysis of its convergence properties is still an active area of 
research~\cite{chen2017extended}. 


A single iteration of our algorithm proceeds by taking the following 
steps for all layers $j=1,\dotsc,l$:
\begin{enumerate}
\item First, $\boldsymbol{w}_j$ is updated by minimizing the Lagrangian after fixing the 
associated auxiliary variable $\boldsymbol{z}_j$ from the previous iteration along with that of 
the previous layer $\boldsymbol{z}_{j-1}$ from the current iteration:
\begin{align}
\boldsymbol{w}_{j}^{[t+1]} & \coloneqq\underset{\boldsymbol{w}_{j}}{\arg\min}\,\mathcal{L}_\rho(\boldsymbol{w}_{j},\boldsymbol{z}_{j-1}^{[t+1]},\boldsymbol{z}_{j}^{[t]},\boldsymbol{\lambda}_{j}^{[t]}) 
\label{eq:admm_w} 
\\[-0.5em]
&
=\left(\mathbf{B}_{j}^{\mathsf{T}}\mathbf{B}_{j}+\rho\mathbf{I}\right)^{-1}(\mathbf{B}_{j}^{\mathsf{T}}\boldsymbol{z}_{j-1}^{[t+1]}+\rho\boldsymbol{z}_{j}^{[t]}-\boldsymbol{\lambda}_{j}^{[t]})
\nonumber 
\end{align}
The solution to this unconstrained least squares problem is found by solving a 
linear system of equations. 

\item Next, $\boldsymbol{z}_j$ is updated by fixing the newly updated $\boldsymbol{w}_j$ 
along with the next layer's coefficients $\boldsymbol{w}_{j+1}$ from the previous iteration:
\begin{align}
\boldsymbol{z}_{j}^{[t+1]} & \coloneqq\underset{\boldsymbol{z}_{j}}{\arg\min}\,\mathcal{L}_\rho(\boldsymbol{w}_{j}^{[t+1]},\boldsymbol{w}_{j+1}^{[t]},\boldsymbol{z}_{j},\boldsymbol{\lambda}_{j}^{[t]})
\label{eq:admm_z}\\[-0.5em]
 &
=\phi_{j}\big(\tfrac{1}{\rho+1}\mathbf{B}_{j+1}\boldsymbol{w}_{j+1}^{[t]}+\tfrac{\rho}{\rho+1}(\boldsymbol{w}_{j}^{[t+1]}+\tfrac{1}{\rho}\boldsymbol{\lambda}_{j}^{[t]})\big)\nonumber\\
\boldsymbol{z}_{l}^{[t+1]}  &
\coloneqq\phi_{j}\big(\boldsymbol{w}_{j}^{[t+1]}+\tfrac{1}{\rho}\boldsymbol{\lambda}_{j}^{[t]}\big)
\nonumber
\end{align}
This is the proximal minimization problem from Eq.~\ref{eq:prox}, so its solution 
is given in closed form via the proximal operator $\boldsymbol{\phi}_j$ associated
with the penalty function $\boldsymbol{\Phi}_j$. For $j\neq l$, its argument is a 
convex combination of the current coefficients $\boldsymbol{w}_j$ and feedback that enforces consistency with the next layer. 

\item Finally, the dual variables $\boldsymbol{\lambda}_j$ are updated with scaled 
constraint violations. 
\begin{equation}
\boldsymbol{\lambda}_{j}^{[t+1]}\coloneqq\boldsymbol{\lambda}_{j}^{[t]}+\rho(\boldsymbol{w}_{j}^{[t+1]}-\boldsymbol{z}_{j}^{[t+1]})
\label{eq:admm_dual}
\end{equation}

\end{enumerate}

This process is then repeated until convergence.
Though not available as a closed-form expression, in the
next section we demonstrate how this algorithm can be posed as a recurrent 
generalization of a feed-forward neural network.

\section{Alternating Direction Neural Networks} \label{sec:adnn}

Our inference algorithm essentially follows the same pattern as a 
deep neural network: for each layer, a learned linear transformation is applied to the 
current output followed by a fixed nonlinear function. Building upon this observation,
we implement it using a recurrent network with standard
layers, thus allowing the model parameters to be learned using backpropagation. 

Recall that the $\boldsymbol{w}_j$ update in Eq.~\ref{eq:admm_w} requires solving
a linear system of equations. While differentiable, this introduces additional computational 
complexity not present in standard neural networks. 
To overcome this, we implicitly assume that the parameters in over-complete layers are
Parseval tight frames, i.e. so that 
$\mathbf{B}_j\mathbf{B}_j^{\mathsf{T}}=\mathbf{I}$.
This property is theoretically advantageous in the field of sparse 
approximation~\cite{casazza2012finite} and has been used as a constraint to encourage 
robustness in deep neural networks~\cite{moustapha2017parseval}. However, in our experiments
we found that it was unnecessary to explicitly enforce this assumption during training; with
appropriate learning rates, backpropagating through our inference algorithm was enough to 
ensure that repeated iterations did not result in diverging sequences of variable updates. 
Thus, under this assumption, we can simplify the update in Eq.~\ref{eq:admm_w} using the Woodbury matrix identity as 
follows: 
\begin{equation}
\boldsymbol{w}_{j}^{[t+1]} \coloneqq \tilde{\boldsymbol{z}}_{j}^{[t]}+\tfrac{1}{\rho+1}\mathbf{B}_{j}^{\mathsf{T}}\big(\boldsymbol{z}_{j-1}^{[t+1]}-\mathbf{B}_{j}\tilde{\boldsymbol{z}}_{j}^{[t]}\big), 
\quad
\tilde{\boldsymbol{z}}_{j}^{[t]} \coloneqq \boldsymbol{z}_{j}^{[t]}-\tfrac{1}{\rho}\boldsymbol{\lambda}_{j}^{[t]}
\label{eq:admm_w_parseval}
\end{equation}

As this only involves simple linear transformations, our ADMM algorithm for solving the 
optimization problem in our inference function $\boldsymbol{f}^{*}$ can be expressed 
as a recurrent neural network that repeatedly iterates until convergence. 
In practice, however, we unroll the network to a fixed number of iterations $T$ for an 
approximation of optimal inference so that  
$\boldsymbol{f}^{[T]}(\boldsymbol{x})\approx\boldsymbol{f}^{*}(\boldsymbol{x})$. Our full algorithm is summarized in Algs.~\ref{alg:feed-forward} and \ref{alg:adnn}.

\begin{figure}[tb]
{
\begin{minipage}[t][][b]{0.34\columnwidth}
\TitleOfAlgo{\centering Feed-Forward} 
\vspace{-0.3em}
\begin{algorithm}[H]
\vspace{0.3em}
	\textbf{Input:} $\boldsymbol{x}$, $\{\mathbf{B}_j,\boldsymbol{b}_j\}$
	
	\textbf{Output:} $\{\boldsymbol{w}_j\}$, $\{\boldsymbol{z}_j\}$
	
	\textbf{Initialize:} $\boldsymbol{z}_0 = \boldsymbol{x}$

	\For{$j=1,\dotsc,l$}{
	\vspace{0.3em}
		\textbf{Pre-activation:} 
		
		\quad$\boldsymbol{w}_j \coloneqq \mathbf{B}_j^{\mathsf{T}}\boldsymbol{z}_{j-1}$
		
		\textbf{Activation:} 
		
		\quad$\boldsymbol{z}_j \coloneqq \phi_j(\boldsymbol{w}_{j} - \boldsymbol{b}_j)$
			}
	\vspace{2.943em}
	\label{alg:feed-forward}
\end{algorithm}
\end{minipage}
\begin{minipage}[t][][b]{0.65\columnwidth}
\TitleOfAlgo{\centering Alternating Direction Neural Network}
\vspace{-0.3em}
\begin{algorithm}[H]
\vspace{0.3em}
	\textbf{Input:} $\boldsymbol{x}$, $\{\mathbf{B}_j,\boldsymbol{b}_j\}$
	
	\textbf{Output:} $\{\boldsymbol{w}_j^{[T]}\}$, $\{\boldsymbol{z}_j^{[T]}\}$
	
	\textbf{Initialize:} $\{\boldsymbol{\lambda}_j^{[0]}\}=\boldsymbol{0}$, $\{\boldsymbol{w}^{[1]}_j,\boldsymbol{z}^{[1]}_j\}$ from Alg.~\ref{alg:feed-forward}
	
	\For{$t=1,\dotsc,T-1$}{
		\For{$j=1,\dotsc,l$}{
			\textbf{Dual:} Update $\boldsymbol{\lambda}_j^{[t]}$ (Eq.~\ref{eq:admm_dual})
			
			\textbf{Pre-activation:} Update $\boldsymbol{w}_{j}^{[t+1]}$ 
			(Eq.~\ref{eq:admm_w_parseval})
			
			\textbf{Activation:} Update $\boldsymbol{z}_{j}^{[t+1]}$ (Eq.~\ref{eq:admm_z})
		}
	}
	\label{alg:adnn}
\end{algorithm}
\end{minipage}
}
\end{figure}

\subsection{Generalization of Feed-Forward Networks}

Given proper initialization of the variables, a single iteration of this algorithm
is identical to a pass through a feed-forward network. Specifically, if we let
$\boldsymbol{\lambda}_{j}^{[0]}=\boldsymbol{0}$ and  
$\boldsymbol{z}_{j}^{[0]}=\mathbf{B}_{j}^{\mathsf{T}}\boldsymbol{z}_{j-1}^{[1]}$,
where we again denote $\boldsymbol{z}_{0}^{[1]}=\boldsymbol{x}$, then 
$\boldsymbol{w}_{j}^{[1]}$ is equivalent to the pre-activation of a neural network
layer:
\begin{equation}
\boldsymbol{w}_{j}^{[1]}\coloneqq\boldsymbol{z}_{j}^{[0]}+\tfrac{1}{\rho+1}\mathbf{B}_{j}^{\mathsf{T}}\big(\boldsymbol{z}_{j-1}^{[1]}-\mathbf{B}_{j}(\mathbf{B}_{j}^{\mathsf{T}}\boldsymbol{z}_{j-1}^{[1]})\big)=\mathbf{B}_{j}^{\mathsf{T}}\boldsymbol{z}_{j-1}^{[1]}
\label{eq:admm_w_init}
\end{equation}

Similarly, if we initialize 
$\boldsymbol{w}_{j+1}^{[0]}=\mathbf{B}_{j+1}^{\mathsf{T}}\boldsymbol{w}_{j}^{[1]}$,  then
$\boldsymbol{z}_{j}^{[1]}$ is equivalent to the corresponding nonlinear activation using the 
proximal operator $\phi_j$: 
\begin{equation}
\boldsymbol{z}_{j}^{[1]}\coloneqq\phi_j\big(\tfrac{1}{\rho+1}\mathbf{B}_{j+1}(\mathbf{B}_{j+1}^{\mathsf{T}}\boldsymbol{w}_{j}^{[1]})+\tfrac{\rho}{\rho+1}\boldsymbol{w}_{j}^{[1]}\big)=\phi_j\big(\boldsymbol{w}_{j}^{[1]}\big)
\label{eq:admm_z_init}
\end{equation}

Thus, one iteration of our inference algorithm is equivalent to the standard feed-forward 
neural network given in Eq.~\ref{eq:feed_forward},  i.e. 
$\boldsymbol{f}^{[1]}(\boldsymbol{x})=\boldsymbol{f}^{\mathrm{DNN}}(\boldsymbol{x})$, where nonlinear activation functions are interpreted as
proximal operators corresponding to the penalties of our DeepCA model. Additional iterations
through the network lead to more accurate inference approximations while explicitly satisfying 
constraints on the latent variables.

\subsection{Learning by Backpropagation}

With DeepCA inference approximated by differentiable ADNNs, the model parameters
can be learned in the same way as standard feed-forward networks. Extending the 
nested component analysis optimization problem from Eq.~\ref{eq:ca_inference}, the 
inference function $\boldsymbol{f}^{[T]}$ can be used as a generalization of feed-forward 
network inference $\boldsymbol{f}^{[1]}$ for  
backpropagation with arbitrary loss functions $L$ that encourage the output to be consistent
with provided supervision $\boldsymbol{y}^{(i)}$, as shown in 
Eq.~\ref{eq:adnn_super} below. Here, only the latent coefficients 
$\boldsymbol{f}_{l}^{[T]}(\boldsymbol{x}^{(i)})$ from the last layer are shown in the loss function, 
but other intermediate outputs $j\neq l$ could also be included.
\begin{equation}
\underset{\{\mathbf{B}_j,\boldsymbol{b}_j\}}{\arg\min}\sum_{i=1}^{n}L\big(\boldsymbol{f}_{l}^{[T]}(\boldsymbol{x}^{(i)}),\,\boldsymbol{y}^{(i)}\big)
\label{eq:adnn_super}
\end{equation}
From an agnostic perspective, an ADNN can 
thus be seen as an end-to-end deep network 
architecture with a particular sequence of linear and nonlinear transformations and tied 
weights. More 
iterations ($T>1$) result in networks with greater effective depth, potentially allowing 
for the representation of more complex nonlinearities. 
However, because the network architecture was derived from an algorithm for inference in our DeepCA model instead 
of arbitrary compositions of parameterized transformations, 
the greater depth requires no additional parameters and serves the very specific
purpose of satisfying constraints on the latent variables while enforcing consistency with
the model parameters. 

\subsection{Theoretical Insights}

In addition to the practical advantages of recurrent ADNNs for constraint satisfaction, 
our DeepCA model provides a useful theoretical tool for better understanding 
traditional neural networks. 
In prior work, Papyan et al. analyzed feed-forward networks as a method for    
approximate inference in a multilayer convolutional sparse coding 
model~\cite{papyan2017convolutional}, which can be seen as a special case of 
DeepCA with overcomplete dictionaries and fixed constraints on the sparsity of the 
coefficients subject to exact reconstructions in each layer. 
Specifically, 
they provide conditions under which the activations of a
feed-forward convolutional network with ReLU nonlinearities approximate the true coefficients 
of their model with bounded error and exact support recovery. While these conditions are 
likely too strict to be satisfied in practice (see Fig.~\ref{fig:explaining_away}, 
for example), the theoretical connection emphasizes the importance of sparsity 
and demonstrates the potential for analyzing deep neural networks from the 
perspective of sparse approximation theory. 

Our more general DeepCA model introduces penalties that relax the requirement of exact 
reconstruction at each 
layer, effectively introducing errors that break up the standard compositional structure 
of neural networks. Importantly, this decoupling of layers allows our original multilayer
inference function in 
Eq.~\ref{eq:deepca} to be expressed as the equivalent shallow learning problem: 
\begin{equation}
\boldsymbol{f}^{*}(\boldsymbol{x})=\underset{\{\boldsymbol{w}_{j}\}}{\arg\min}\,\frac{1}{2}\left\lVert \left[\begin{array}{c}
\boldsymbol{x}\\
\mathbf{0}\\[-0.3em]
\vdots\\
\mathbf{0}
\end{array}\right]-\left[\begin{array}{cccc}
\mathbf{B}_{1} & \mathbf{0} &  & \mathbf{0}\\[-0.6em]
-\mathbf{I} & \mathbf{B}_{2} & \ddots\\[-0.5em]
 & \ddots & \ddots & \mathbf{0}\\[-0.1em]
\mathbf{0} &  & -\mathbf{I} & \mathbf{B}_{l}
\end{array}\right]\left[\begin{array}{c}
\boldsymbol{w}_{1}\\
\boldsymbol{w}_{2}\\[-0.2em]
\vdots\\[-0.2em]
\boldsymbol{w}_{l}
\end{array}\right]\right\rVert _{2}^{2}+\sum_{j=1}^{l}\Phi_{j}(\boldsymbol{w}_{j})
\label{eq:shallow}
\end{equation}
With nonnegative $\ell_1$ penalties corresponding to 
biased ReLU activation functions, this is simply an augmented, higher-dimensional  
nonnegative sparse coding problem in which the dictionary
is constrained to have a particular block structure that relates to the architecture
of the corresponding neural network. This suggests that  
nonlinear function composition may not be necessary for effective 
deep learning, but instead the implicit structure that it enforces. 

This connection to shallow learning allows for the direct application of results from the field of sparse 
approximation theory. For example, despite being able to exactly 
reconstruct any datapoint, if overcomplete dictionaries satisfy certain incoherence properties, 
then the sparsest set of reconstruction coefficients may actually be unique~\cite{donoho2003optimally}. 
Uniqueness is an important characteristic of learned representations that are able to 
memorize data, a prominent feature in many effective deep architectures~\cite{arpit2017closer}.
While standard shallow models are typically incapable of satisfying the properties required for uniqueness in 
high-dimensional datasets like those common in the field of computer vision, 
DeepCA suggests that the increased capacity of deep networks can be explained 
by the structure imposed on the augmented dictionaries of 
Eq.~\ref{eq:shallow}. Higher-dimensionality allows for richer unique representations with 
a greater number of non-zero elements while the block structure reduces the number of
free parameters that need to be learned.  
Uniqueness guarantees can also apply to solutions of sparse nonnegative least squares 
problems even without explicitly enforcing sparsity~\cite{bruckstein2008uniqueness}, which 
may help to explain the success of deep networks without explicit bias 
terms~\cite{huang2017densely}. More broadly, we believe that DeepCA opens up a fruitful
new direction for future research towards 
the principled design of neural network architectures that
optimize capacity for sparse representation by enforcing implicit structure on the 
augmented dictionaries of their associated shallow models.

\begin{figure*}[t]
\centering
\subfloat[Decoder Error]{
\includegraphics[width=0.23\textwidth]{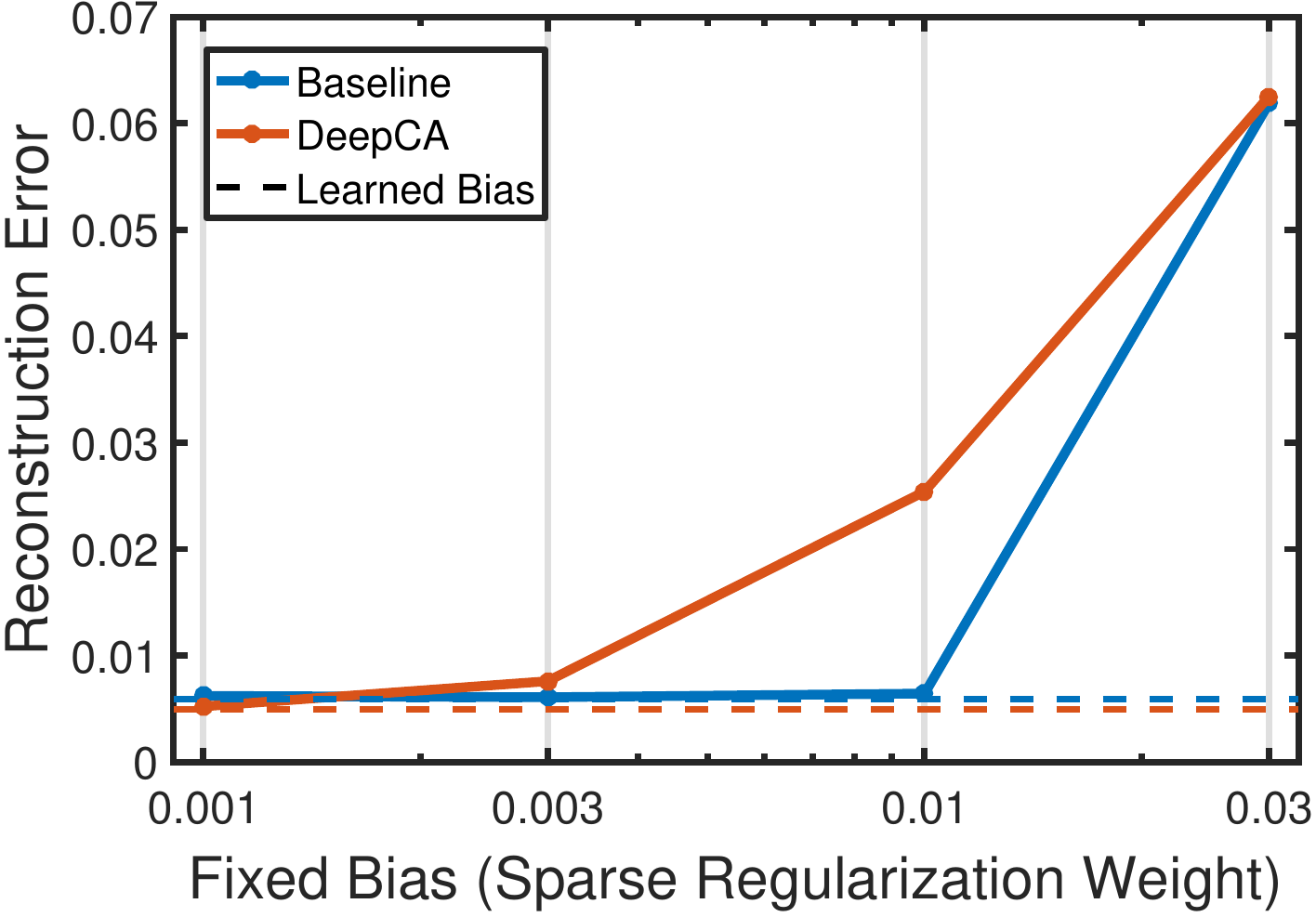}
}
\rulesep
\subfloat[Layer 1 Sparsity]{
\includegraphics[width=0.23\textwidth]{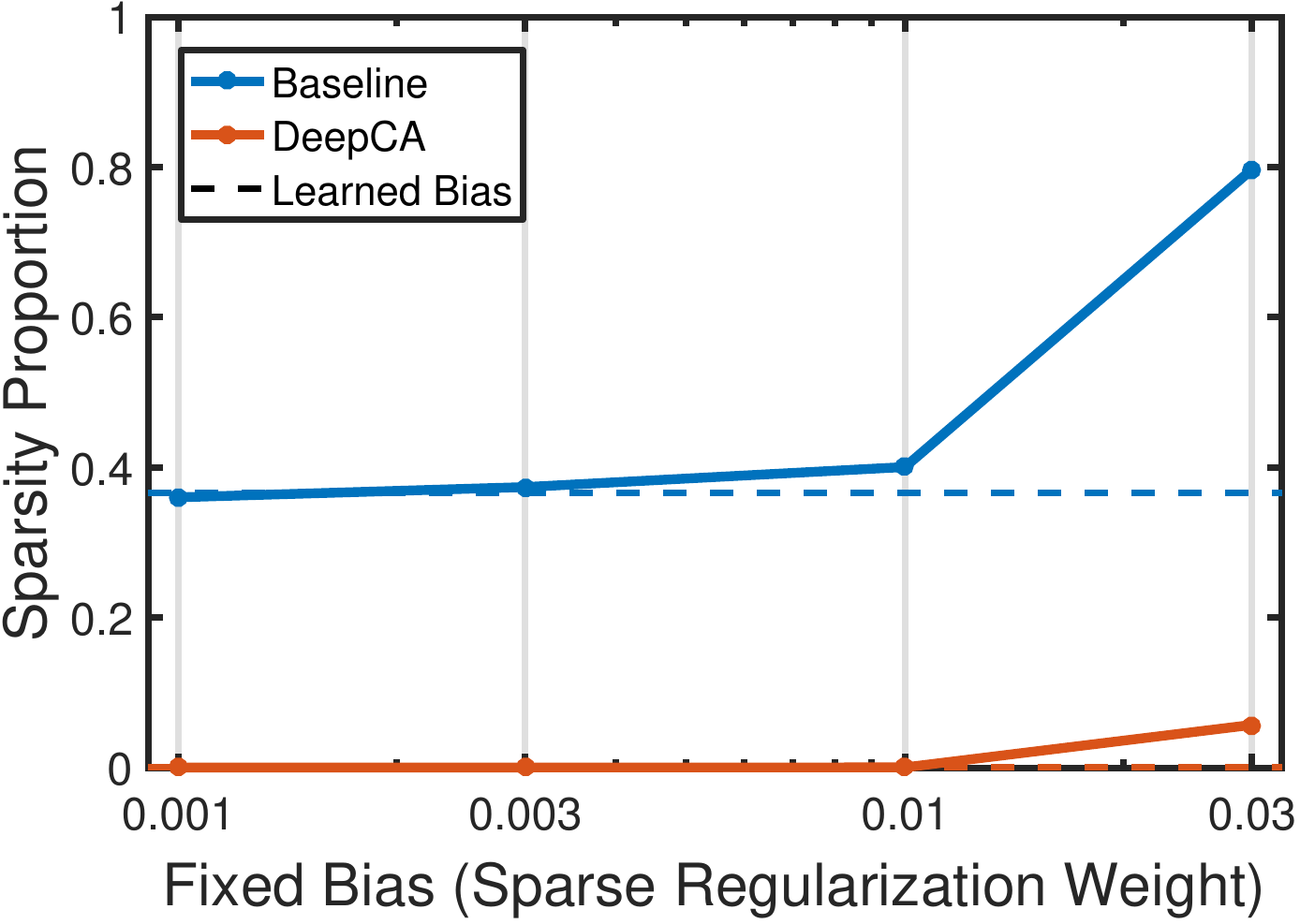}
}
\subfloat[Layer 2 Sparsity]{
\includegraphics[width=0.23\textwidth]{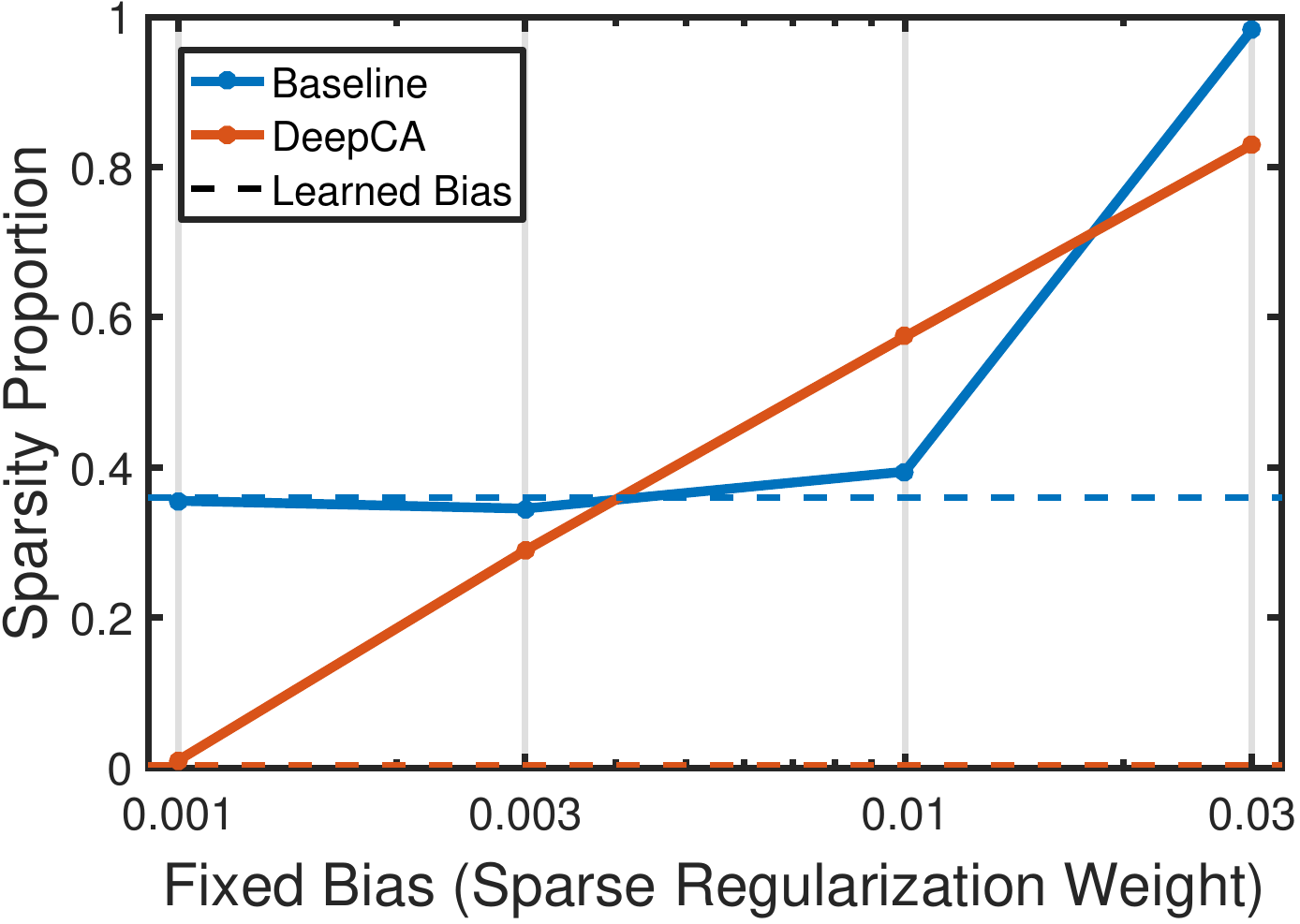}
}
\subfloat[Layer 3 Sparsity]{
\includegraphics[width=0.23\textwidth]{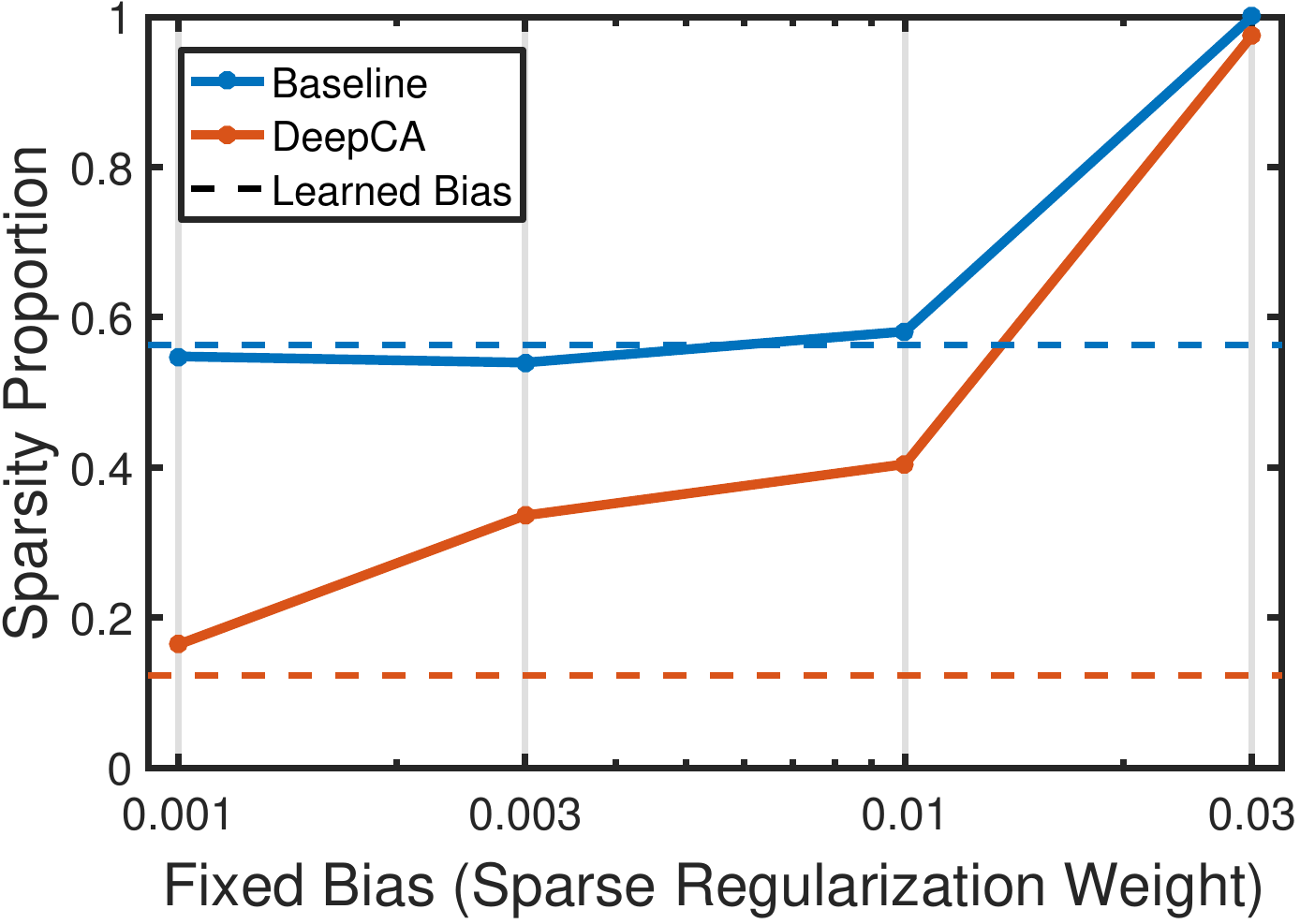}
}

\caption{
A demonstration of the effects of fixed (solid lines) and 
learnable (dotted lines) bias parameters on the reconstruction error (a) and activation 
sparsity (b-d) comparing feed forward networks (blue) with DeepCA (red). All models consist of three layers each with 512 components. Due to the conditional 
dependence provided by recurrent feedback, DeepCA learns to better control the sparsity level 
in order improve reconstruction error. As $\ell_1$ regularization weights, the biases converge 
towards zero resulting in denser activations and higher network capacity for reconstruction.
}
\label{fig:cifar_unsuper_bias}
\end{figure*}

\begin{figure*}[tb]
\centering
\hspace{\fill}
\subfloat[Training Error]{
\includegraphics[height=2.8cm]{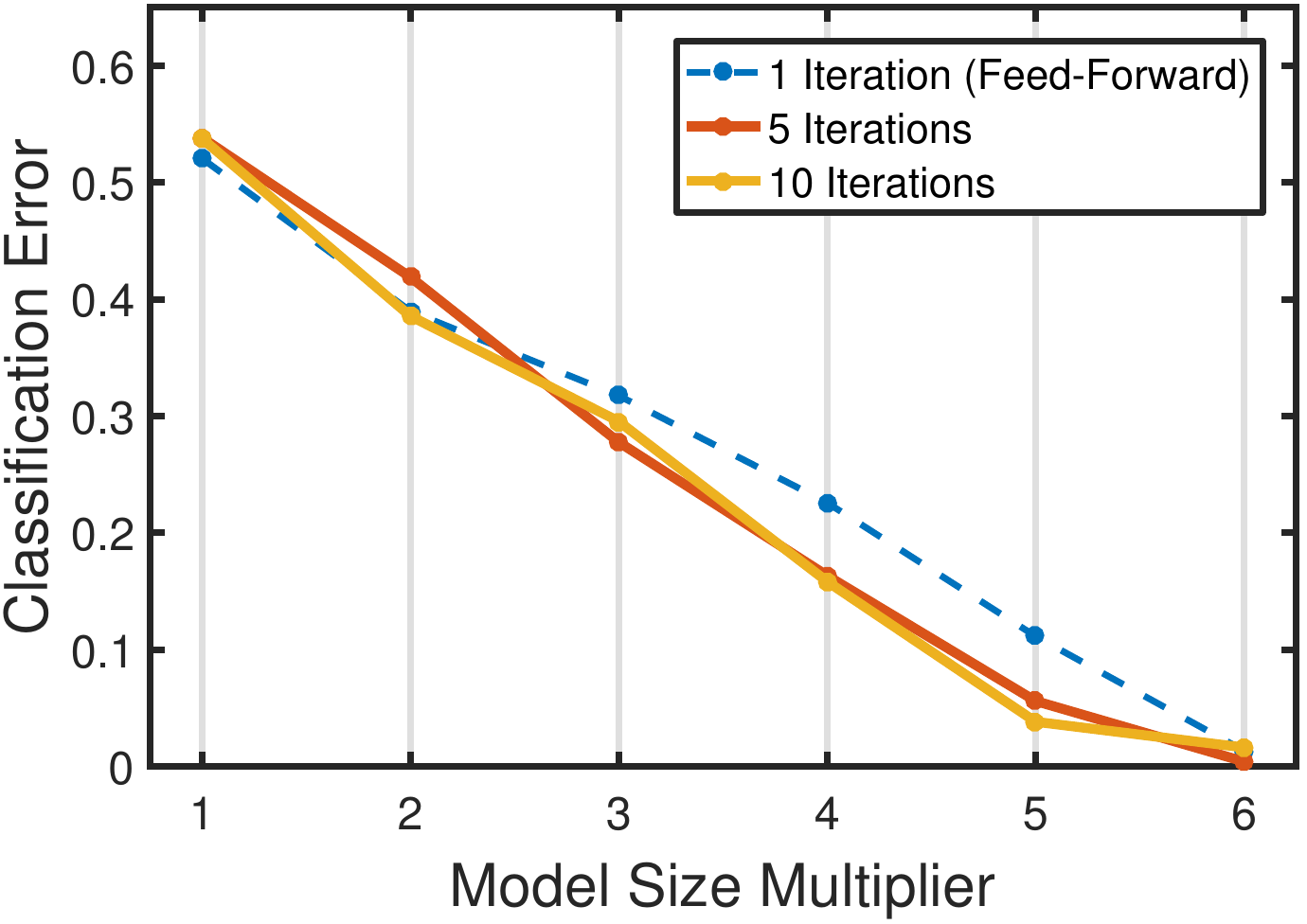}
}
\hspace{\fill}
\subfloat[Testing Error]{
\includegraphics[height=2.8cm]{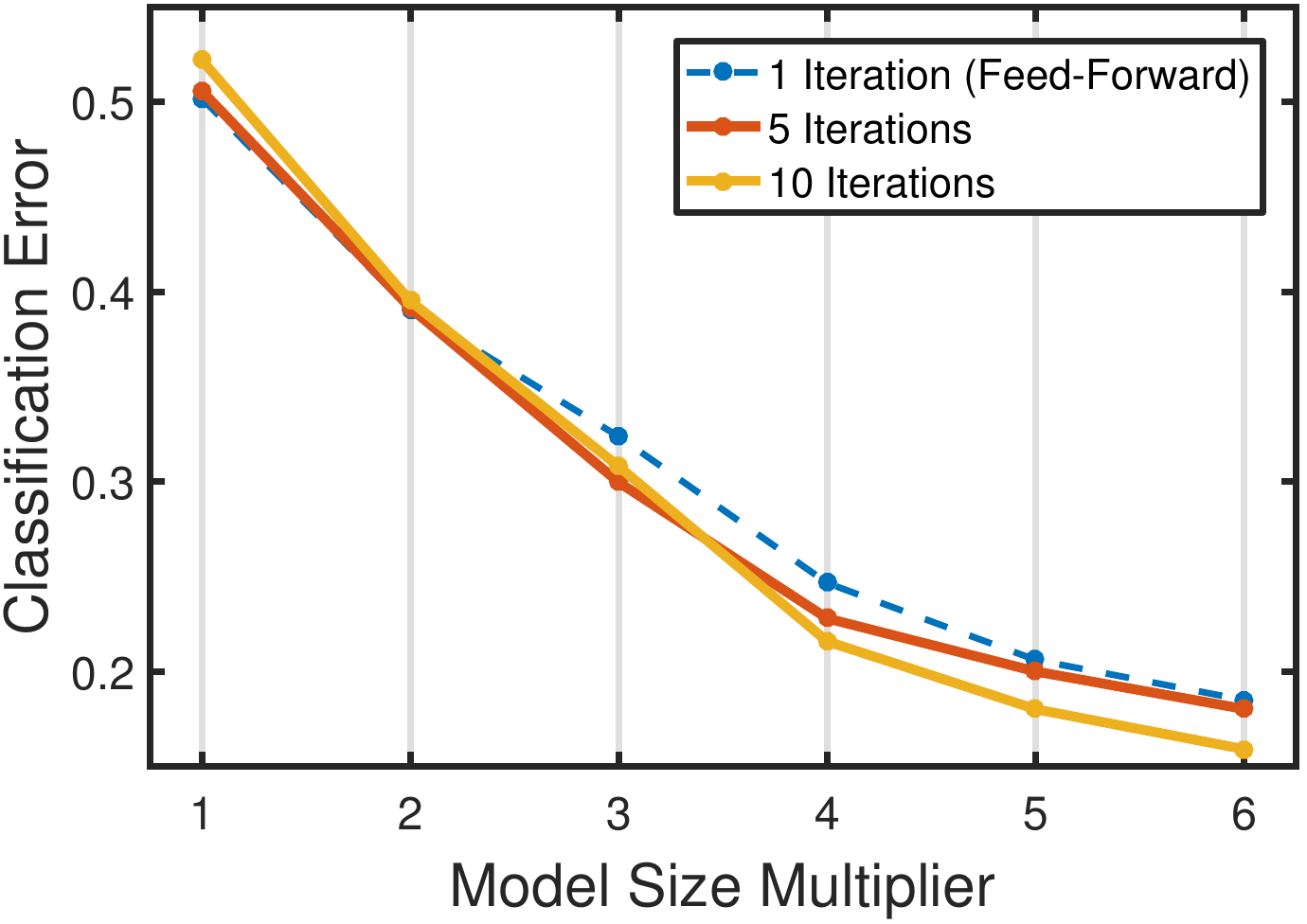}
}
\hspace{\fill}
\subfloat[Optimization]{
\includegraphics[height=2.8cm]{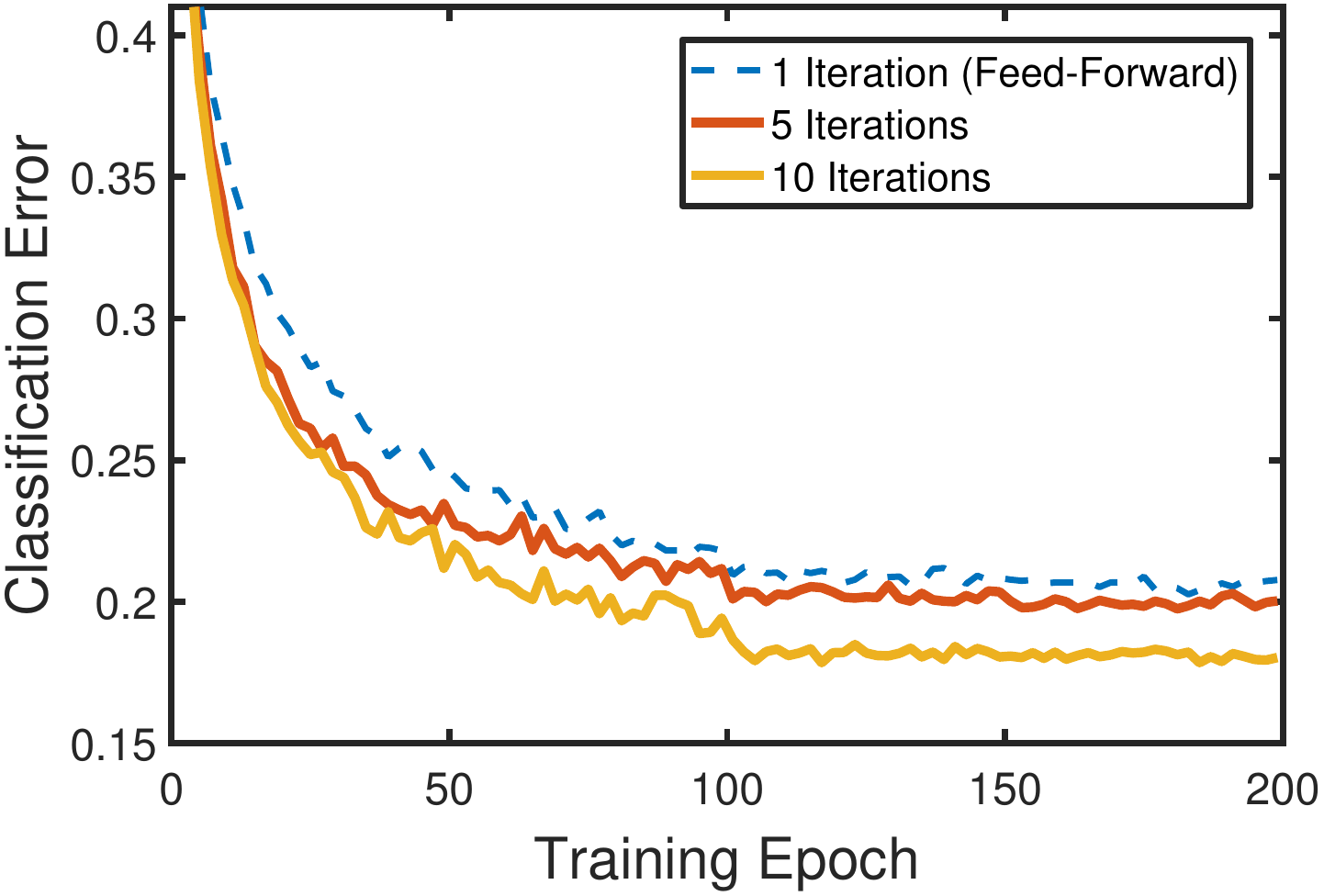}
}
\hspace{\fill}
\caption{
The effect of increasing model size on training (a) and testing (b) classification 
error, demonstrating consistently improved performance of ADNNs over
feed-forward networks, especially in larger models. The base model consists of two $3\times 3$, 
2-strided convolutional layers followed by one fully-connected layer with 4, 8, and 16 
components respectively. Also shown are is the classification error throughout training (c). 
}
\label{fig:cifar_super}
\end{figure*}

\begin{figure}[t]
\centering
\subfloat[Training]{
\includegraphics[width=0.23\columnwidth]{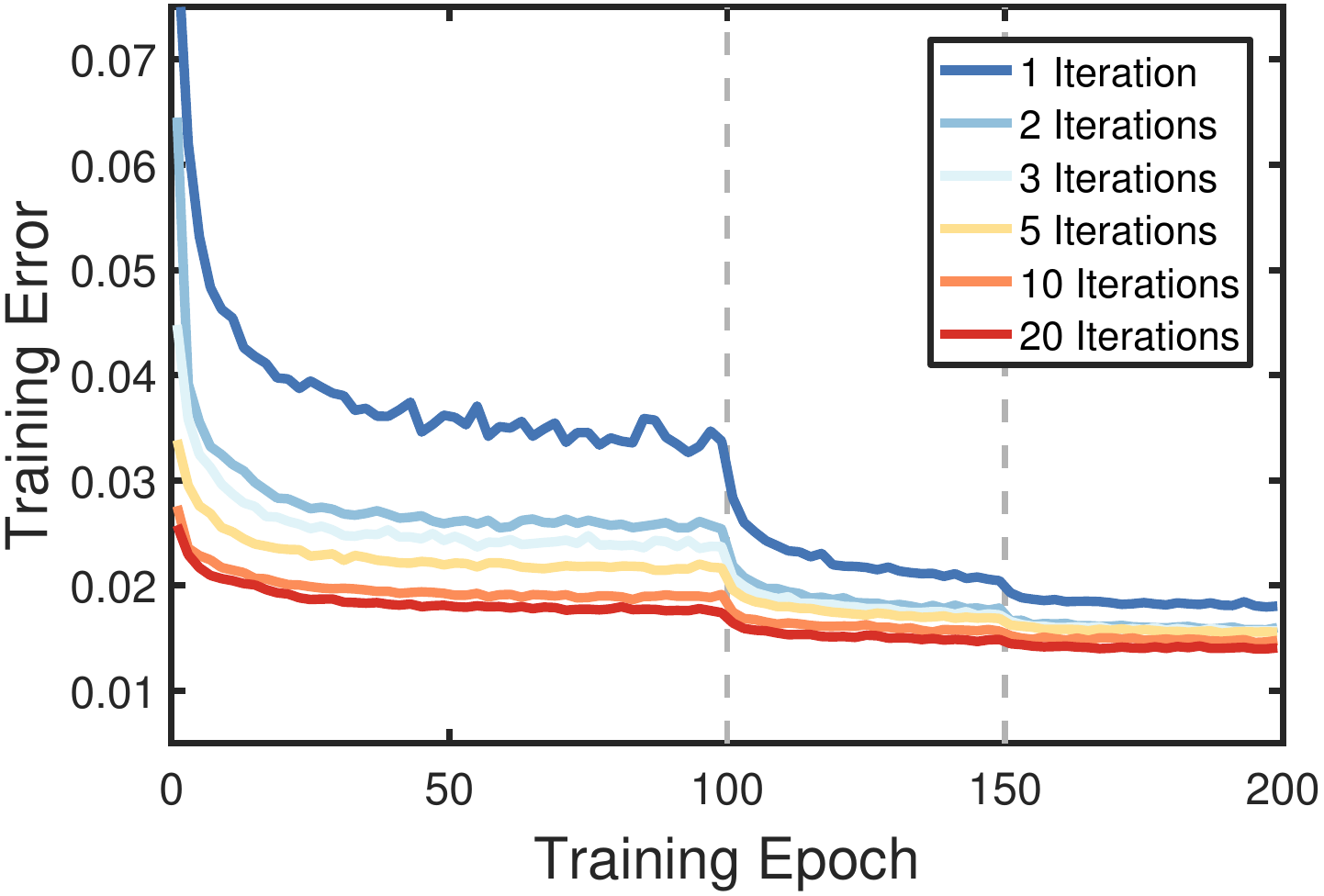}
}
\subfloat[Testing]{
\includegraphics[width=0.23\columnwidth]{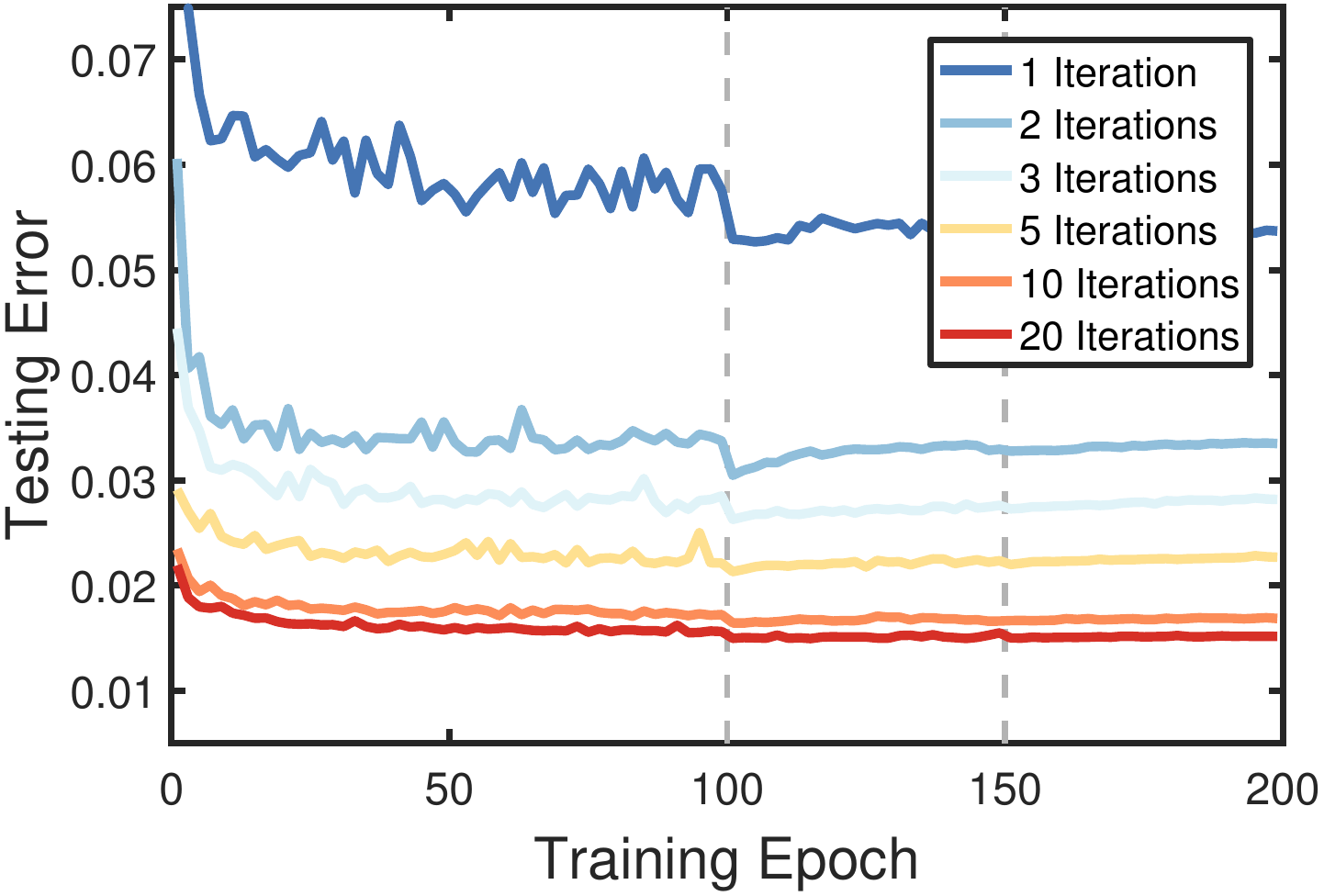}
}
\rulesep
\subfloat[Train Error]{
\includegraphics[width=0.23\columnwidth]{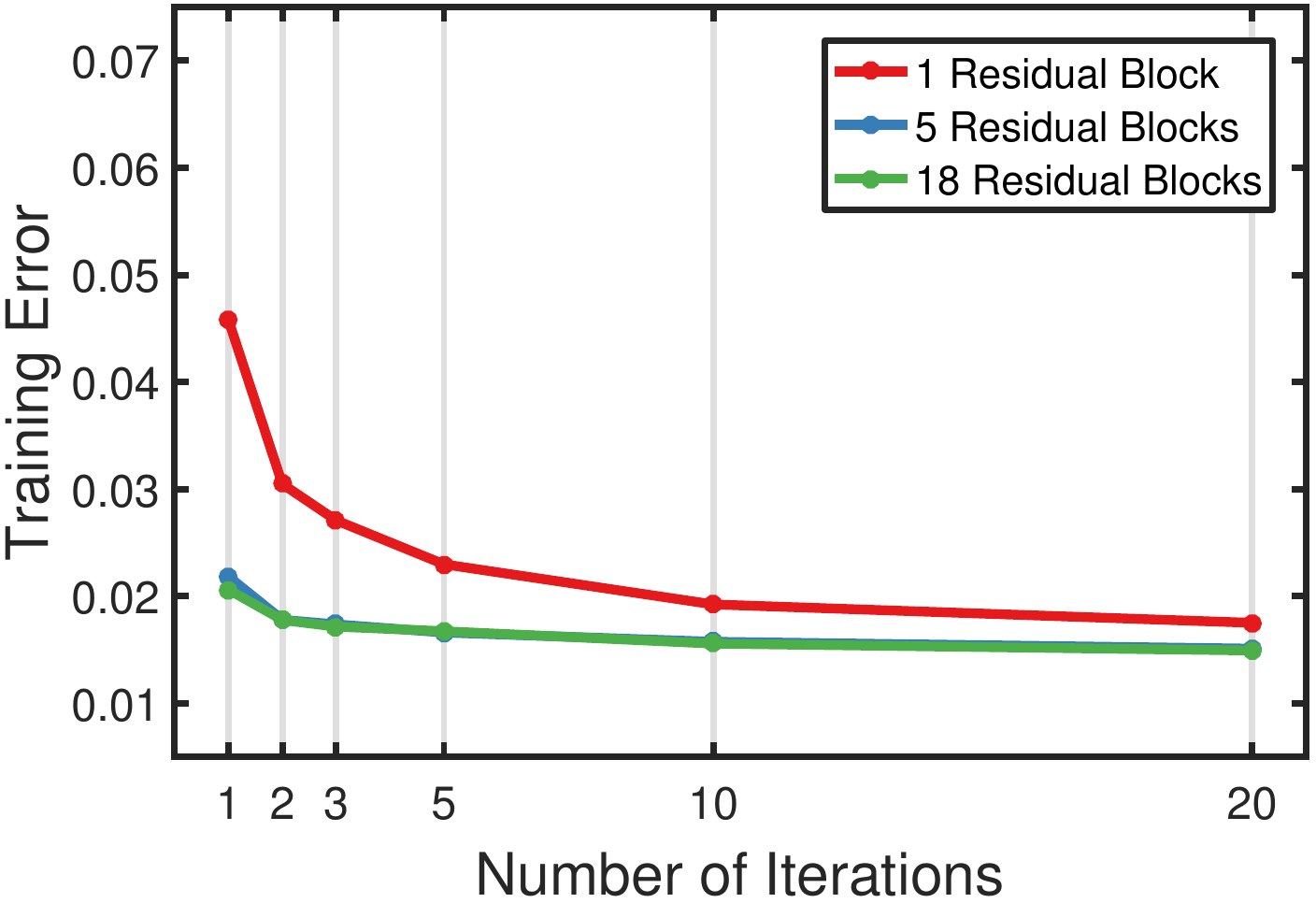}
}
\subfloat[Test Error]{
\includegraphics[width=0.23\columnwidth]{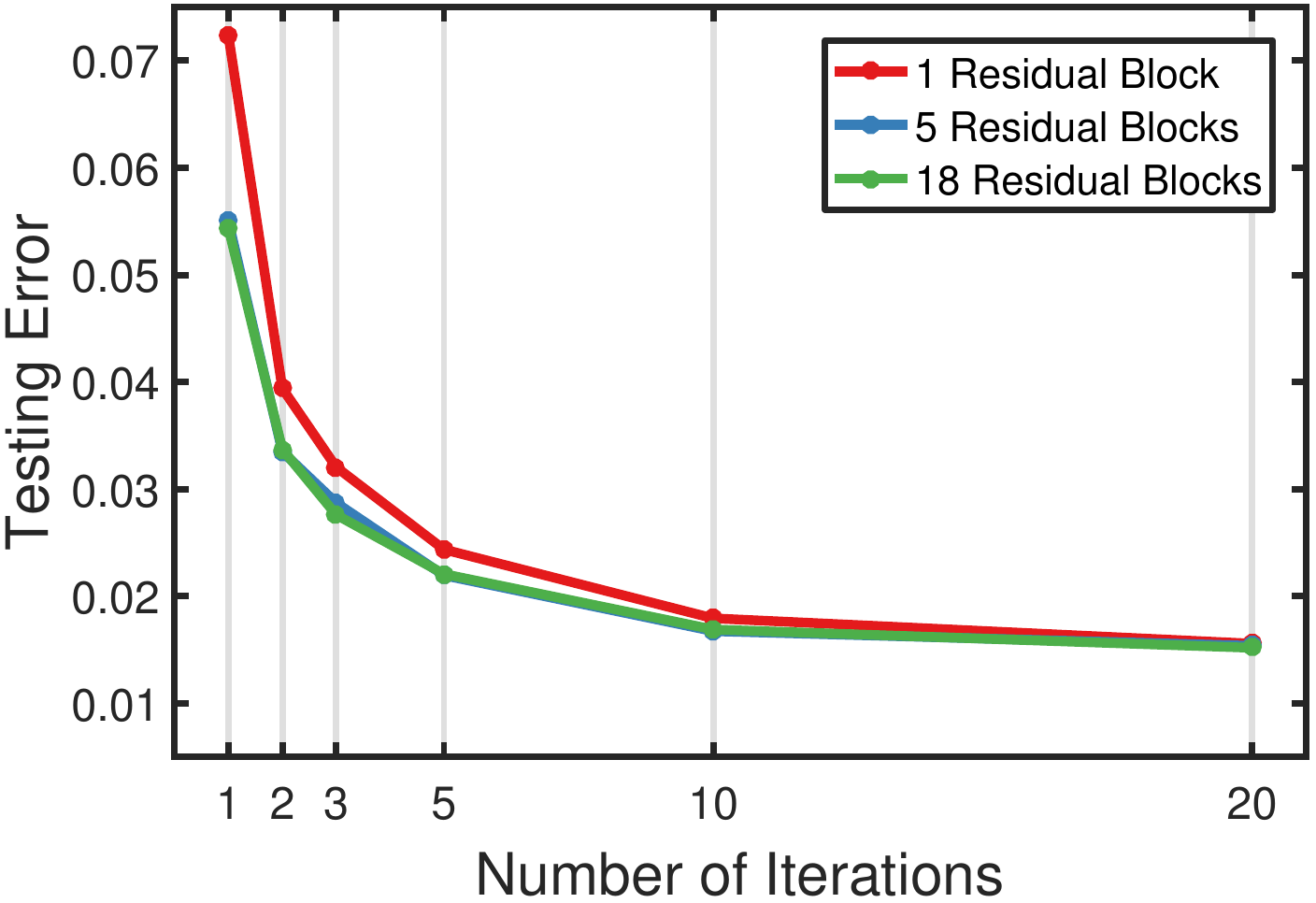}
}

\caption{
Quantitative results demonstrating the improved generalization performance of ADNN inference. 
The training (a) and testing (b) reconstruction errors throughout optimization show that more 
iterations ($T>1$) substantially reduce convergence time and give much lower error on 
held-out test data. With a sufficiently large number of iterations, even lower-capacity models 
with encoders consisting of fewer residual blocks all achieve nearly the same level of 
performance with small discrepancies between training (c) and testing (d) errors.
}
\label{fig:generalization}
\end{figure}

\begin{figure}[tb]
\centering
\captionsetup[subfigure]{labelformat=empty}

\subfloat[]{\makebox[0.9em]{\small(a)} \makebox[0.9em][l]{\rotatebox[origin=c]{90}{\small{Input}}}\makebox[1.2em][l]{\rotatebox[origin=c]{90}{\small{Image}}}}
\subfloat[]{\vcenteredinclude{width=0.10\columnwidth}{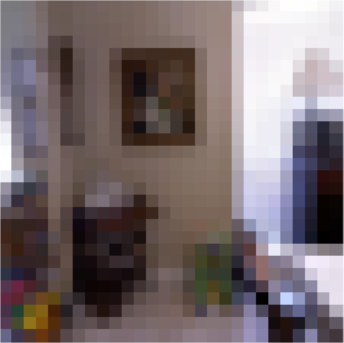}}\hspace{-0.2em}
\subfloat[]{\vcenteredinclude{width=0.10\columnwidth}{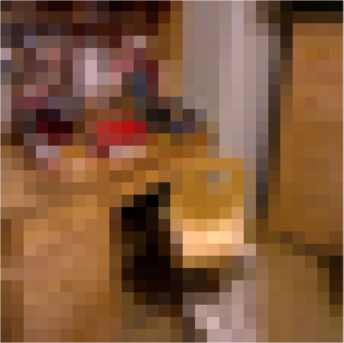}}\hspace{-0.2em}
\subfloat[]{\vcenteredinclude{width=0.10\columnwidth}{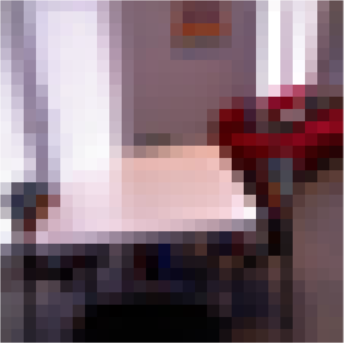}}\hspace{-0.2em}
\subfloat[]{\vcenteredinclude{width=0.10\columnwidth}{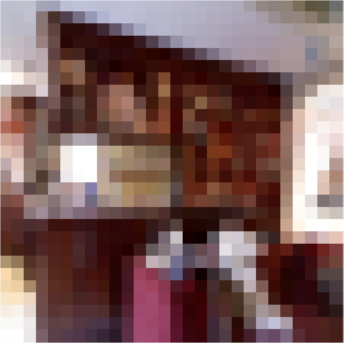}}\hspace{-0.2em}
\subfloat[]{\vcenteredinclude{width=0.10\columnwidth}{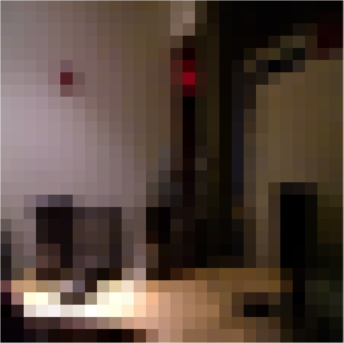}}\hspace{-0.2em}
\subfloat[]{\vcenteredinclude{width=0.10\columnwidth}{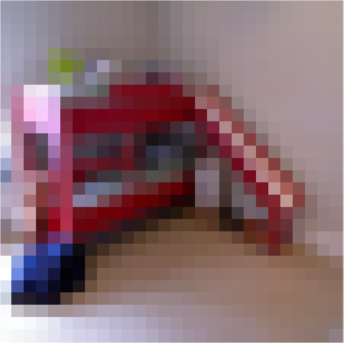}}\hspace{-0.2em}
\subfloat[]{\vcenteredinclude{width=0.10\columnwidth}{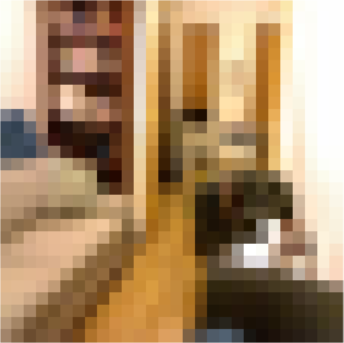}}\hspace{-0.2em}
\subfloat[]{\vcenteredinclude{width=0.10\columnwidth}{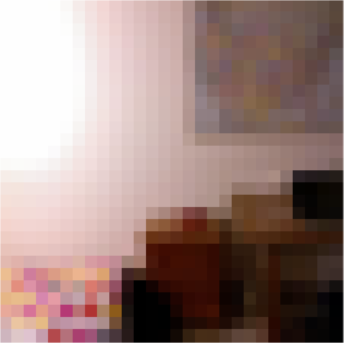}}\hspace{-0.2em}
\subfloat[]{\vcenteredinclude{width=0.10\columnwidth}{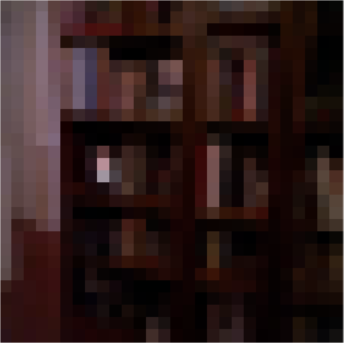}}

\vspace{-2.3em}
\subfloat[]{\makebox[0.9em]{\small(b)} \makebox[0.9em][l]{\rotatebox[origin=c]{90}{\small{Baseline}}}\makebox[1.2em][l]{\rotatebox[origin=c]{90}{\small{($T=1$)}}}}
\subfloat[]{\vcenteredinclude{width=0.10\columnwidth}{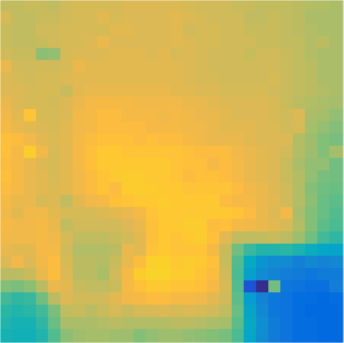}}\hspace{-0.2em}
\subfloat[]{\vcenteredinclude{width=0.10\columnwidth}{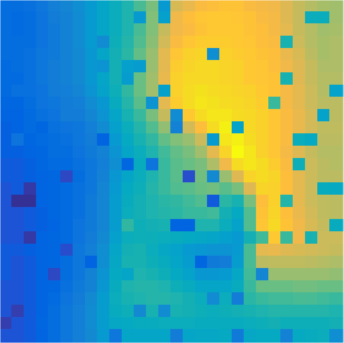}}\hspace{-0.2em}
\subfloat[]{\vcenteredinclude{width=0.10\columnwidth}{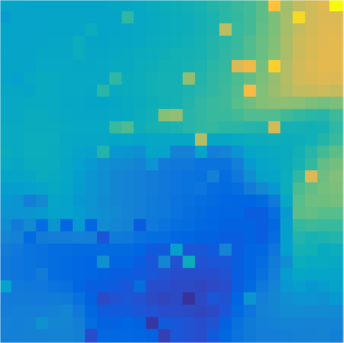}}\hspace{-0.2em}
\subfloat[]{\vcenteredinclude{width=0.10\columnwidth}{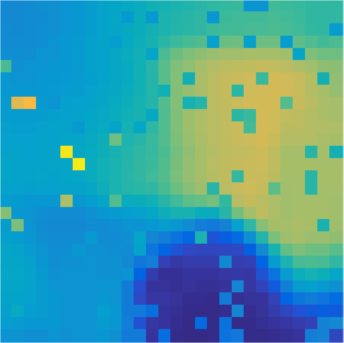}}\hspace{-0.2em}
\subfloat[]{\vcenteredinclude{width=0.10\columnwidth}{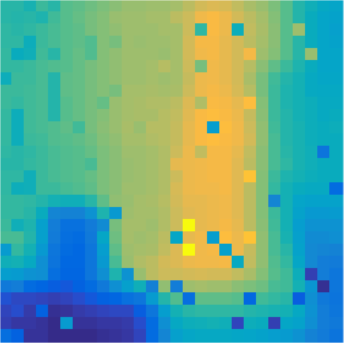}}\hspace{-0.2em}
\subfloat[]{\vcenteredinclude{width=0.10\columnwidth}{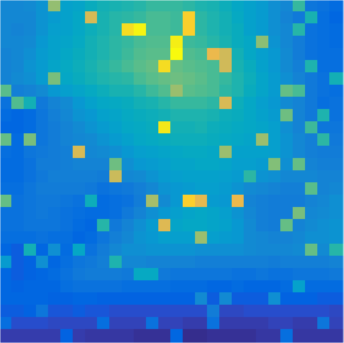}}\hspace{-0.2em}
\subfloat[]{\vcenteredinclude{width=0.10\columnwidth}{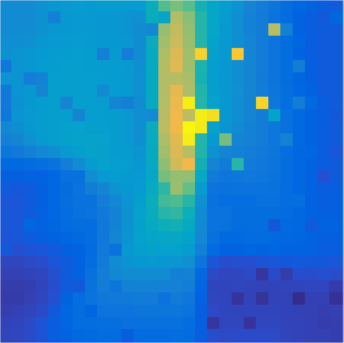}}\hspace{-0.2em}
\subfloat[]{\vcenteredinclude{width=0.10\columnwidth}{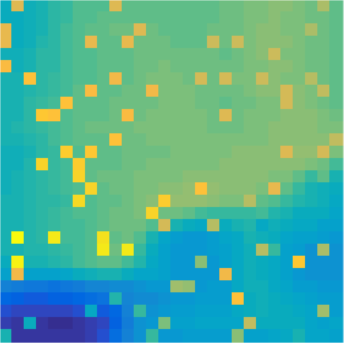}}\hspace{-0.2em}
\subfloat[]{\vcenteredinclude{width=0.10\columnwidth}{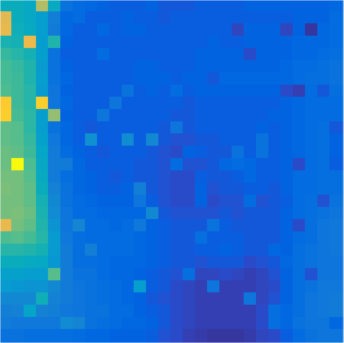}}

\vspace{-2.3em}
\subfloat[]{\makebox[0.9em]{\small(c)} \makebox[0.9em][l]{\rotatebox[origin=c]{90}{\small{\bfseries{ADNN}}}}\makebox[1.2em][l]{\rotatebox[origin=c]{90}{\small{($\boldsymbol{T=20}$)}}}}
\subfloat[]{\vcenteredinclude{width=0.10\columnwidth}{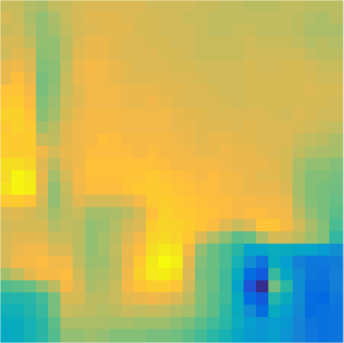}}\hspace{-0.2em}
\subfloat[]{\vcenteredinclude{width=0.10\columnwidth}{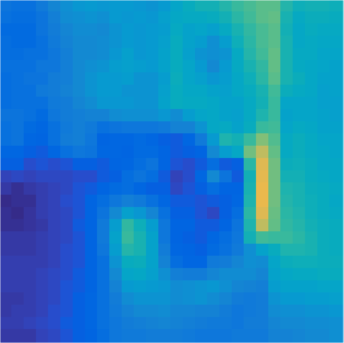}}\hspace{-0.2em}
\subfloat[]{\vcenteredinclude{width=0.10\columnwidth}{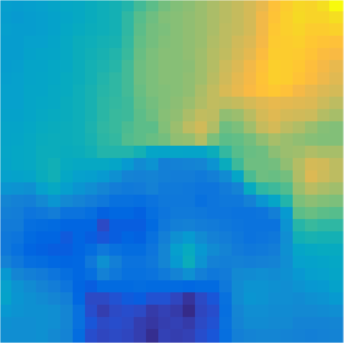}}\hspace{-0.2em}
\subfloat[]{\vcenteredinclude{width=0.10\columnwidth}{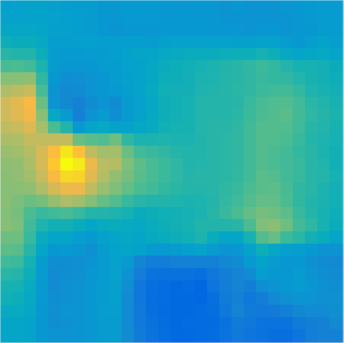}}\hspace{-0.2em}
\subfloat[]{\vcenteredinclude{width=0.10\columnwidth}{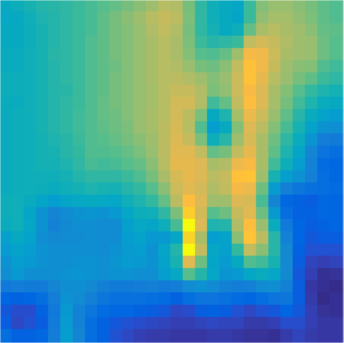}}\hspace{-0.2em}
\subfloat[]{\vcenteredinclude{width=0.10\columnwidth}{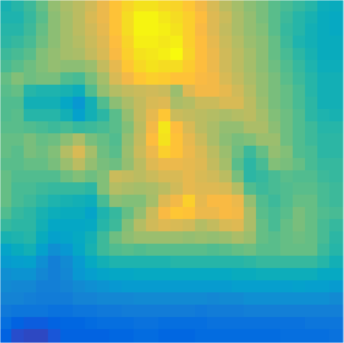}}\hspace{-0.2em}
\subfloat[]{\vcenteredinclude{width=0.10\columnwidth}{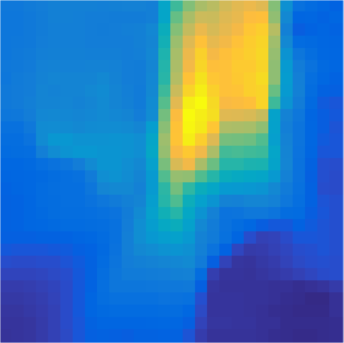}}\hspace{-0.2em}
\subfloat[]{\vcenteredinclude{width=0.10\columnwidth}{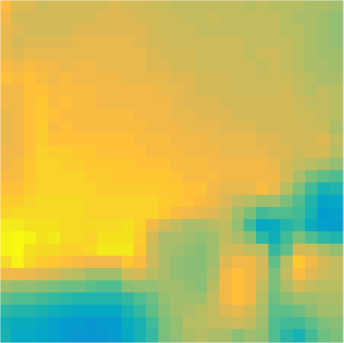}}\hspace{-0.2em}
\subfloat[]{\vcenteredinclude{width=0.10\columnwidth}{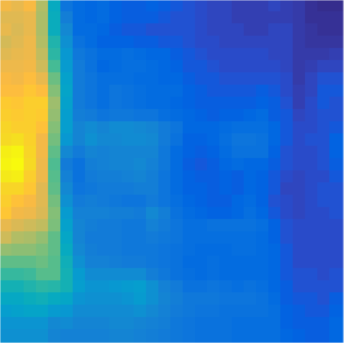}}

\vspace{-2.3em}
\subfloat[]{\makebox[0.9em]{\small(d)} \makebox[0.9em][l]{\rotatebox[origin=c]{90}{\small{Ground}}}\makebox[1.2em][l]{\rotatebox[origin=c]{90}{\small{Truth}}}}
\subfloat[(i)]{\vcenteredinclude{width=0.10\columnwidth}{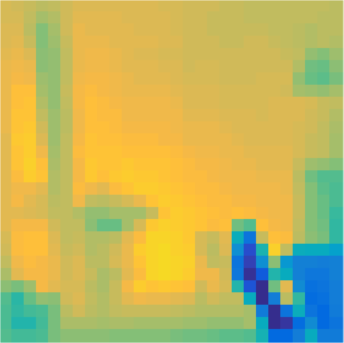}}\hspace{-0.2em}
\subfloat[(ii)]{\vcenteredinclude{width=0.10\columnwidth}{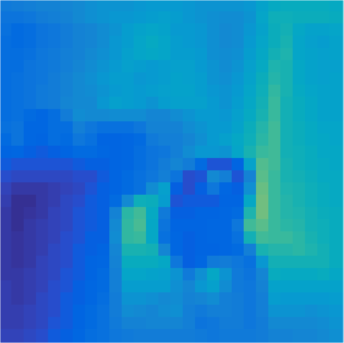}}\hspace{-0.2em}
\subfloat[(iii)]{\vcenteredinclude{width=0.10\columnwidth}{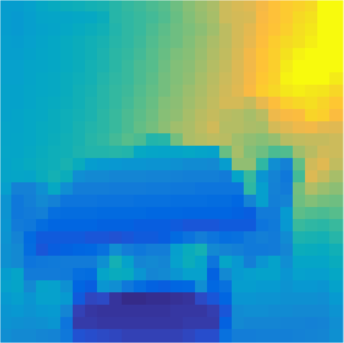}}\hspace{-0.2em}
\subfloat[(iv)]{\vcenteredinclude{width=0.10\columnwidth}{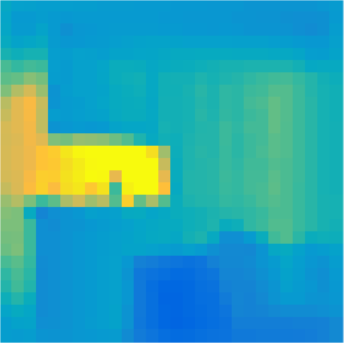}}\hspace{-0.2em}
\subfloat[(v)]{\vcenteredinclude{width=0.10\columnwidth}{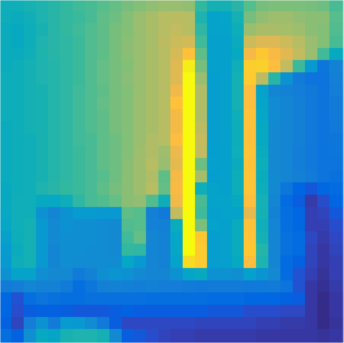}}\hspace{-0.2em}
\subfloat[(vi)]{\vcenteredinclude{width=0.10\columnwidth}{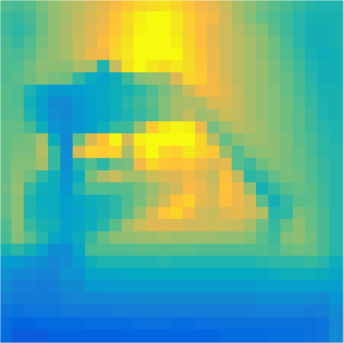}}\hspace{-0.2em}
\subfloat[(vii)]{\vcenteredinclude{width=0.10\columnwidth}{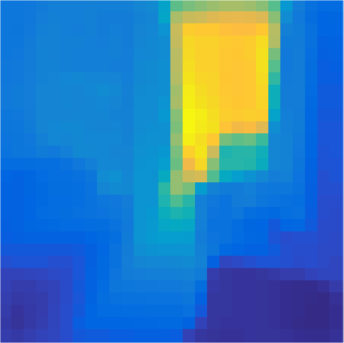}}\hspace{-0.2em}
\subfloat[(viii)]{\vcenteredinclude{width=0.10\columnwidth}{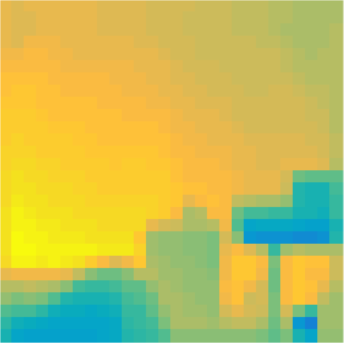}}\hspace{-0.2em}
\subfloat[(ix)]{\vcenteredinclude{width=0.10\columnwidth}{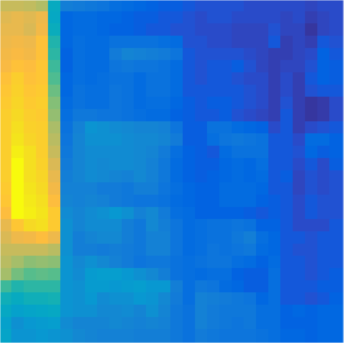}}

\subfloat[]{\makebox[0.9em]{\small(a)} \makebox[0.9em][l]{\rotatebox[origin=c]{90}{\small{Input}}}\makebox[1.2em][l]{\rotatebox[origin=c]{90}{\small{Image}}}}
\subfloat[]{\vcenteredinclude{width=0.10\columnwidth}{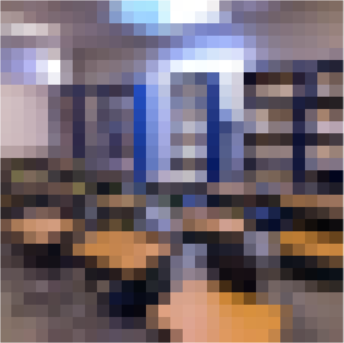}}\hspace{-0.2em}
\subfloat[]{\vcenteredinclude{width=0.10\columnwidth}{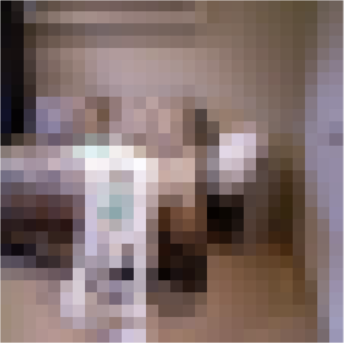}}\hspace{-0.2em}
\subfloat[]{\vcenteredinclude{width=0.10\columnwidth}{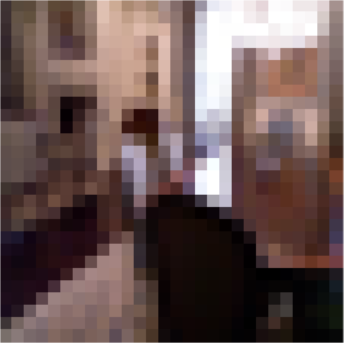}}\hspace{-0.2em}
\subfloat[]{\vcenteredinclude{width=0.10\columnwidth}{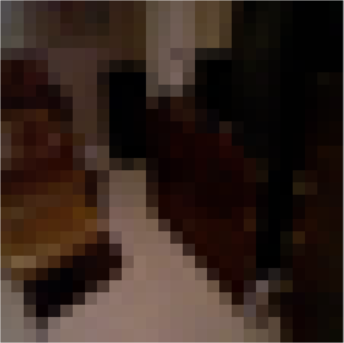}}\hspace{-0.2em}
\subfloat[]{\vcenteredinclude{width=0.10\columnwidth}{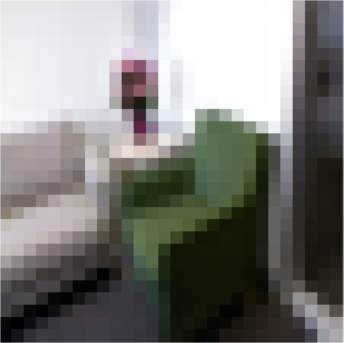}}\hspace{-0.2em}
\subfloat[]{\vcenteredinclude{width=0.10\columnwidth}{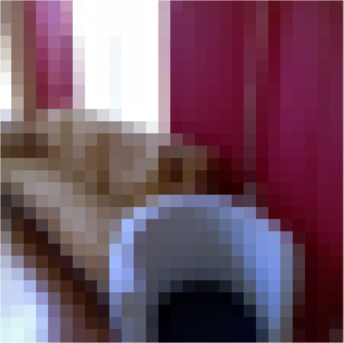}}\hspace{-0.2em}
\subfloat[]{\vcenteredinclude{width=0.10\columnwidth}{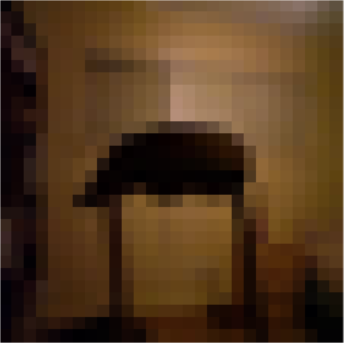}}\hspace{-0.2em}
\subfloat[]{\vcenteredinclude{width=0.10\columnwidth}{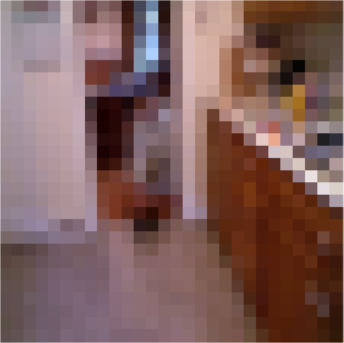}}\hspace{-0.2em}
\subfloat[]{\vcenteredinclude{width=0.10\columnwidth}{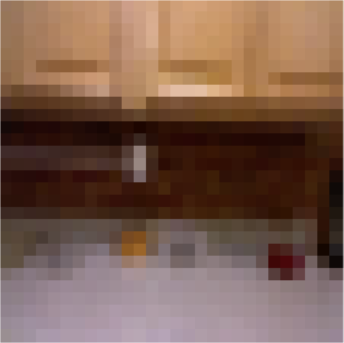}}

\vspace{-2.3em}
\subfloat[]{\makebox[0.9em]{\small(b)} \makebox[0.9em][l]{\rotatebox[origin=c]{90}{\small{Baseline}}}\makebox[1.2em][l]{\rotatebox[origin=c]{90}{\small{($T=1$)}}}}
\subfloat[]{\vcenteredinclude{width=0.10\columnwidth}{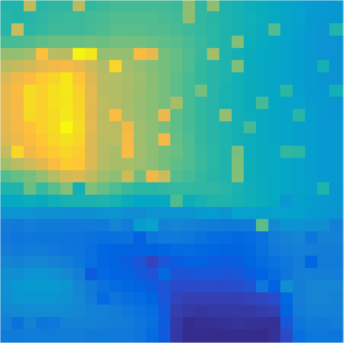}}\hspace{-0.2em}
\subfloat[]{\vcenteredinclude{width=0.10\columnwidth}{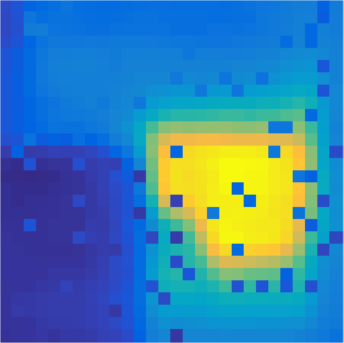}}\hspace{-0.2em}
\subfloat[]{\vcenteredinclude{width=0.10\columnwidth}{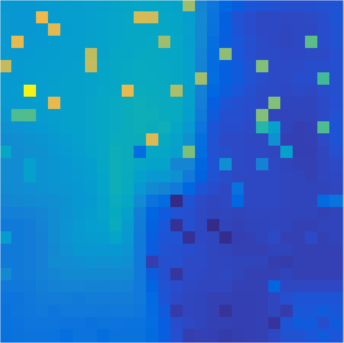}}\hspace{-0.2em}
\subfloat[]{\vcenteredinclude{width=0.10\columnwidth}{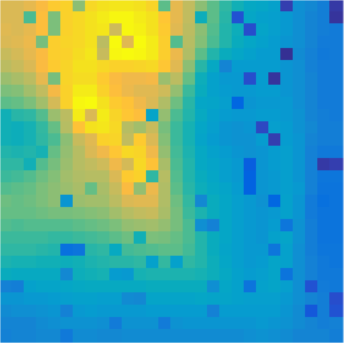}}\hspace{-0.2em}
\subfloat[]{\vcenteredinclude{width=0.10\columnwidth}{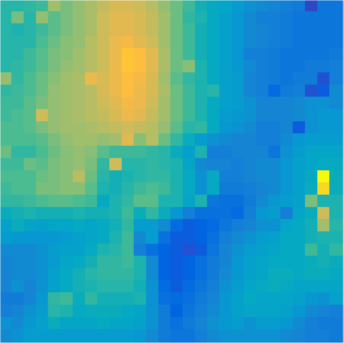}}\hspace{-0.2em}
\subfloat[]{\vcenteredinclude{width=0.10\columnwidth}{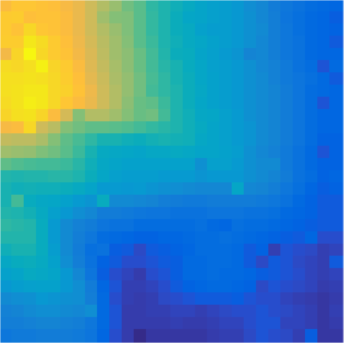}}\hspace{-0.2em}
\subfloat[]{\vcenteredinclude{width=0.10\columnwidth}{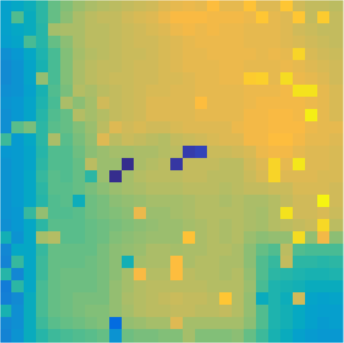}}\hspace{-0.2em}
\subfloat[]{\vcenteredinclude{width=0.10\columnwidth}{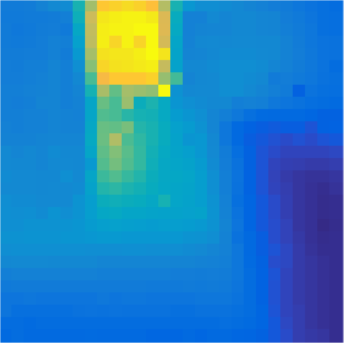}}\hspace{-0.2em}
\subfloat[]{\vcenteredinclude{width=0.10\columnwidth}{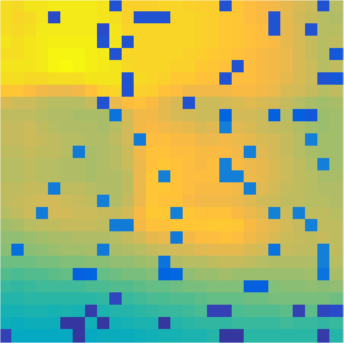}}

\vspace{-2.3em}
\subfloat[]{\makebox[0.9em]{\small(c)} \makebox[0.9em][l]{\rotatebox[origin=c]{90}{\small{\bfseries{ADNN}}}}\makebox[1.2em][l]{\rotatebox[origin=c]{90}{\small{($\boldsymbol{T=20}$)}}}}
\subfloat[]{\vcenteredinclude{width=0.10\columnwidth}{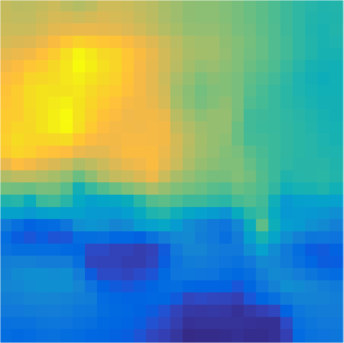}}\hspace{-0.2em}
\subfloat[]{\vcenteredinclude{width=0.10\columnwidth}{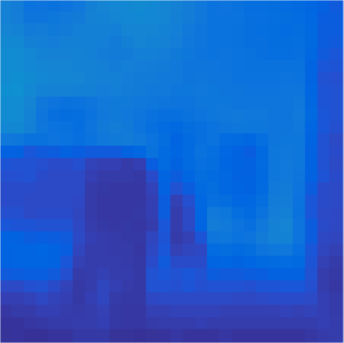}}\hspace{-0.2em}
\subfloat[]{\vcenteredinclude{width=0.10\columnwidth}{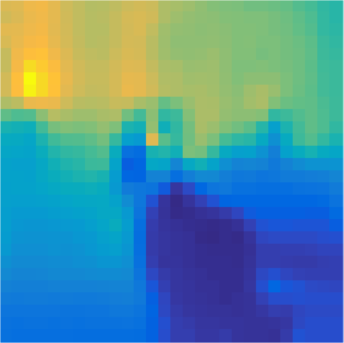}}\hspace{-0.2em}
\subfloat[]{\vcenteredinclude{width=0.10\columnwidth}{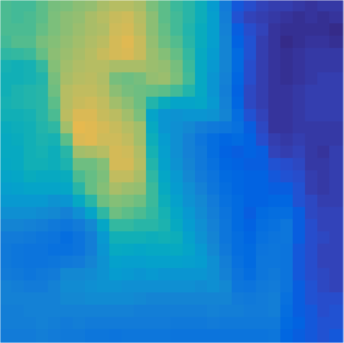}}\hspace{-0.2em}
\subfloat[]{\vcenteredinclude{width=0.10\columnwidth}{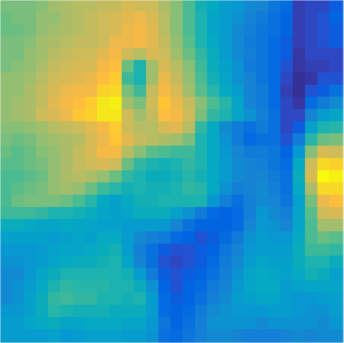}}\hspace{-0.2em}
\subfloat[]{\vcenteredinclude{width=0.10\columnwidth}{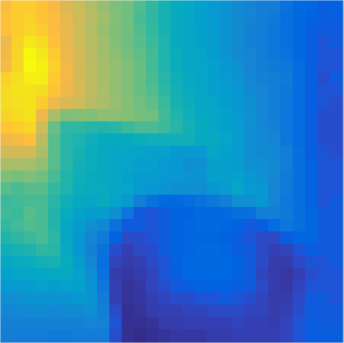}}\hspace{-0.2em}
\subfloat[]{\vcenteredinclude{width=0.10\columnwidth}{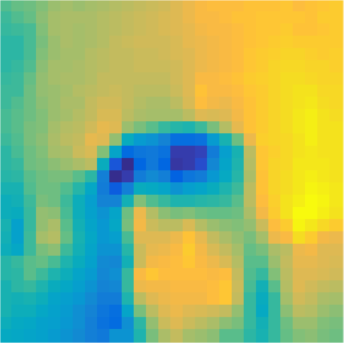}}\hspace{-0.2em}
\subfloat[]{\vcenteredinclude{width=0.10\columnwidth}{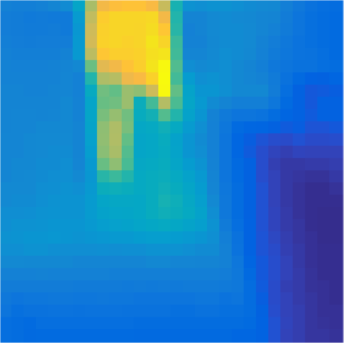}}\hspace{-0.2em}
\subfloat[]{\vcenteredinclude{width=0.10\columnwidth}{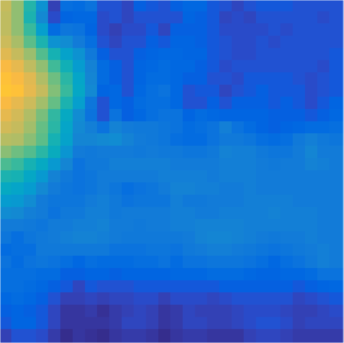}}

\vspace{-2.3em}
\subfloat[]{\makebox[0.9em]{\small(d)} \makebox[0.9em][l]{\rotatebox[origin=c]{90}{\small{Ground}}}\makebox[1.2em][l]{\rotatebox[origin=c]{90}{\small{Truth}}}}
\subfloat[(x)]{\vcenteredinclude{width=0.10\columnwidth}{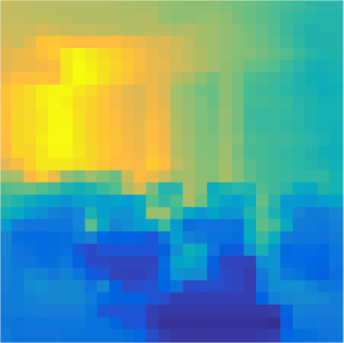}}\hspace{-0.2em}
\subfloat[(xi)]{\vcenteredinclude{width=0.10\columnwidth}{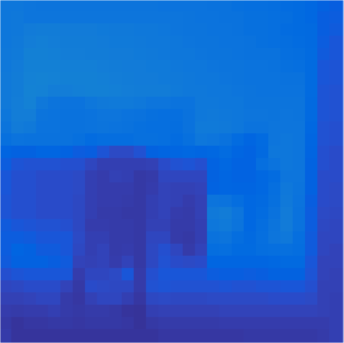}}\hspace{-0.2em}
\subfloat[(xii)]{\vcenteredinclude{width=0.10\columnwidth}{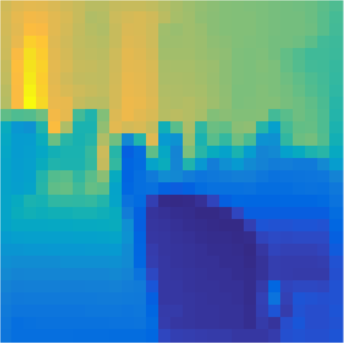}}\hspace{-0.2em}
\subfloat[(xiii)]{\vcenteredinclude{width=0.10\columnwidth}{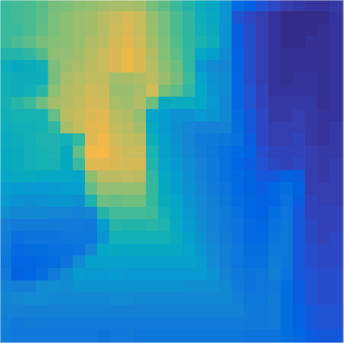}}\hspace{-0.2em}
\subfloat[(xiv)]{\vcenteredinclude{width=0.10\columnwidth}{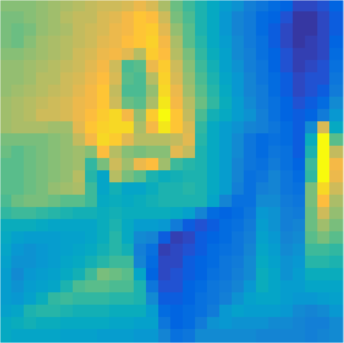}}\hspace{-0.2em}
\subfloat[(xv)]{\vcenteredinclude{width=0.10\columnwidth}{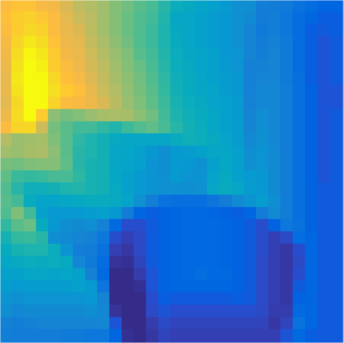}}\hspace{-0.2em}
\subfloat[(xvi)]{\vcenteredinclude{width=0.10\columnwidth}{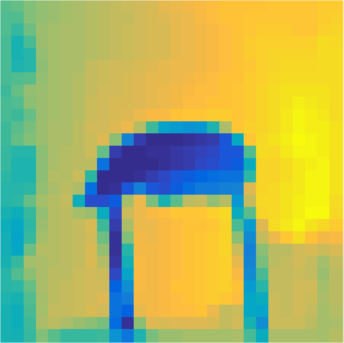}}\hspace{-0.2em}
\subfloat[(xvii)]{\vcenteredinclude{width=0.10\columnwidth}{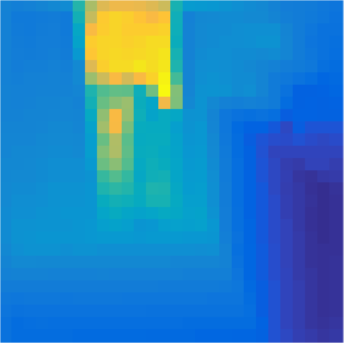}}\hspace{-0.2em}
\subfloat[(xviii)]{\vcenteredinclude{width=0.10\columnwidth}{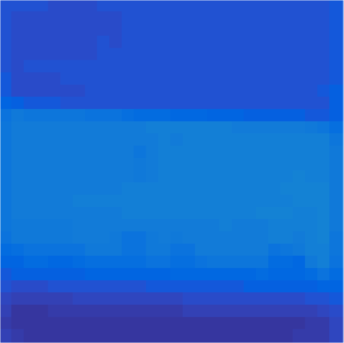}}

\caption{Qualitative depth prediction results given a single image (a) 
and a sparse set of known depth values as input. Outputs of the baseline feed-forward model 
(b) are inconsistent with the constraints as evidenced by unrealistic 
discontinuities. An ADNN with $T=20$ iterations (c) learns to enforce the 
constraints, resolving ambiguities for more detailed predictions that better agree with ground 
truth depth maps (d). Depending on the difficulty, additional iterations 
may have little effect on the output (xvii) or be insufficient to consistently integrate
the known constraint values (xviii).}
\label{fig:examples}
\end{figure}

\section{Experimental Results} \label{sec:experiments}

In this section, we demonstrate some practical advantages of more accurate inference approximations in our
DeepCA model using recurrent ADNNs over feed-forward networks. 
Even without additional prior knowledge, standard convolutional networks with ReLU activation 
functions still benefit from additional recurrent iterations as demonstrated by consistent 
improvements in both supervised and unsupervised tasks on the CIFAR-10 
dataset~\cite{krizhevsky2009learning}.
Specifically, with an unsupervised $\ell_2$ reconstruction loss function, 
Fig.~\ref{fig:cifar_unsuper_bias} shows that
the conditional dependence between features provided by additional iterations allows for better
sparsity control, resulting in higher network capacity through denser activations and 
lower reconstruction error. This suggests that recurrent feedback allows
ADNNs to learn richer representation spaces by explicitly penalizing activation sparsity. 
With a supervised classification loss function, backpropagating through inference allows biases 
to be used for adaptive regularization, providing more freedom in modulating activation sparsity 
to increase model capacity by ensuring the uniqueness of representations across semantic categories. 
This results in improved classification 
performance as shown in Fig.~\ref{fig:cifar_super}, especially for wider models with more 
components per layer.

While these experiments emphasize the importance of sparsity in deep networks and
justify our DeepCA model formulation, the effectiveness of 
feed-forward soft thresholding as an approximation of explicit $\ell_1$ regularization
limits the amount of additional capacity that can be achieved with more iterations. 
As such, ADNNs provide much greater performance gains when prior 
knowledge is available in the form of constraints that \emph{cannot} be effectively 
approximated by feed-forward nonlinearities. This is exemplified by our application of 
output-constrained single-image depth prediction where simple feed-forward correction 
of the known depth values results in inconsistent discontinuities. We demonstrate this with the 
NYU-Depth V2 dataset~\cite{silberman2012nyudepth}, from which we sample 60k 
training images and 500 testing images from held-out scenes. To facilitate faster 
training, we resize the images to $28\times 28$ and then randomly sample 10\% of the ground truth depth 
values to simulate known measurements. Following~\cite{ma2017sparse}, our model architecture uses
a ResNet encoder for feature extraction of the image concatenated with the known depth 
values as an additional input channel. This is followed by an ADNN decoder 
composed of three transposed convolution upsampling layers with biased ReLU nonlinearites in
the first two layers and a constraint correction proximal operator in the last layer. 
Fig.~\ref{fig:generalization} shows the mean absolute prediction errors of this model 
with increasing numbers of iterations and different encoder sizes. While all models have
similar prediction error on training data, ADNNs with more iterations achieve 
significantly improved generalization performance, reducing the test error of the feed-forward 
baseline by over 72\% from 0.054 to 0.015 with 20 iterations even with low-capacity encoders. 
Qualitative visualizations in Fig.~\ref{fig:examples} show that these improvements result from 
consistent constraint satisfaction that serves to resolve depth ambiguities.

\section{Conclusion}

DeepCA is a novel deep model formulation that extends shallow component analysis techniques to 
increase representational capacity. Unlike feed-forward networks, intermediate network 
activations are interpreted as latent reconstruction coefficients to be inferred 
using an iterative constrained optimization algorithm. This is implemented using recurrent ADNNs, 
which allow the model 
parameters to be learned with arbitrary loss functions. In addition, they provide a tool
for consistently integrating prior knowledge in the form of constraints or regularization 
penalties. Due to its close 
relationship to feed-forward networks, which are equivalent to one iteration of this 
algorithm with proximal operators replacing nonlinear activation functions, 
DeepCA also provides a novel theoretical perspective from which to interpret deep 
networks. This suggests the use of sparse approximation theory as tool for analyzing and 
designing network architectures that optimize the capacity for learning unique representations of data. 
\clearpage
\bibliographystyle{splncs}
\bibliography{refs}
\end{document}